\documentclass[journal]{IEEEtran}

\newcommand{\subparagraph}{}

\usepackage[compact]{titlesec}
\usepackage{graphicx}
\usepackage{subcaption}
\usepackage{dsfont}
\usepackage{xcolor}
\usepackage{mathrsfs}
\usepackage{centernot}
\usepackage{mathtools}
\usepackage{amsmath,amsfonts,amssymb,amsthm}
\usepackage{algorithm}
\usepackage{algorithmic}
\usepackage{booktabs}
\usepackage{soul}
\usepackage{cite}

\titlespacing*{\paragraph}
{0pt}{3.25ex plus 1ex minus .2ex}{1.5ex plus .2ex}
\renewcommand{\baselinestretch}{0.98}

\ifCLASSINFOpdf
\else
\fi

\newcommand{\argmax}{\mathop{\mathrm{argmax}}\limits}   % ASdeL
\newcommand*\mystrut[1]{\vrule width0pt height0pt depth#1\relax}

\begin{document}
%!TEX encoding = UTF-8 Unicode
\title{Spatio-Temporal Point Processes with Attention\\for Traffic Congestion Event Modeling}

\author{Shixiang~Zhu,
        Ruyi~Ding,
        Minghe~Zhang,
        Pascal~Van~Hentenryck,
        and~Yao~Xie}% <-this % stops a space
\maketitle

\begin{abstract}
%We present a novel attention-based, mutually exciting spatio-temporal point process model, a versatile framework for mutually dependent modeling traffic congestion events over the road network. We consider multi-modal data by combining traffic counts data from traffic sensor networks with police reports, which contain two types of triggering mechanisms for subsequent congestion events. To capture the non-homogeneous temporal dependence of the event on the past, we introduce a novel attention-based approach for the point process model. To precisely capture the directional spatial dependence induced by the road network, we incorporate such structure by adapting similar ideas used for the ``tail-up'' model, which was previously used in the classical spatial statistics context. We demonstrate the superior performance of our approach compared to the state-of-the-art for both synthetic and real data. 
We present a novel framework for modeling traffic congestion events over road networks. Using multi-modal data by combining count data from traffic sensors with police reports that report traffic incidents, we aim to capture two types of triggering effect for congestion events. Current traffic congestion at one location may cause future congestion over the road network, and traffic incidents may cause spread traffic congestion. To model the non-homogeneous temporal dependence of the event on the past, we use a novel attention-based mechanism based on neural networks embedding for point processes. To incorporate the directional spatial dependence induced by the road network, we adapt the ``tail-up'' model from the context of spatial statistics to the traffic network setting. We demonstrate our approach's superior performance compared to the state-of-the-art methods for both synthetic and real data.

\end{abstract}

\begin{IEEEkeywords}
traffic congestion, attention, point process.
\end{IEEEkeywords}

% \fi
\IEEEpeerreviewmaketitle

% \vspace{-.1in}
\section{Introduction}
\label{sec:intro}

Traffic congestion modeling is critical to modern transportation applications, such as route guidance or traffic network planning. For example, in Atlanta, which has over half a million daily commuters, reducing congestion is a top priority. The city spends millions of dollars on traffic-reducing measures, including toll lanes and high-capacity transport \cite{DavidWickert2020}. However, modeling the complex traffic dynamics and predicting traffic congestion events in real-time is vital but has remained extremely challenging. Indeed, traffic modeling is inherently intricate because of the complex spatio-temporal dynamics and the fact that congestion also stems from responses to real-time events such as traffic accidents. As a result, understanding and predicting congestion events can help cities cope with traffic more efficiently and plan future urban development.

Traffic sensors distributed over highway and road networks are widely deployed. They are key technology enablers that provide a unique opportunity to understand the traffic dynamic and congestion. Traffic sensor data reports traffic counts, i.e.,  the number of cars passing through per unit of time. These traffic counts are exploited by most existing works (reviewed in the section below), but traditional approaches do not model traffic events and incidents, which are fundamentally different in nature. An essential feature of traffic congestion modeling is the ability to capture {\it triggering effects}. For example, when congestion occurs, the effect will propagate, and subsequent congestion is typically more likely to happen along the affected road or highway. Moreover, other types of events with police intervention, such as response to traffic accidents, may also be related to traffic congestion. Such events logged into the police reports provide an additional data source useful for modeling and predicting traffic congestion. 

In this paper, we aim to capture these two types of events and their triggering effects. Hawkes processes (also called self-exciting point processes) are a popular model for modeling such a triggering effect, and they have been successfully used in many different applications (see \cite{Reinhart2017} for a review). A Hawkes process models the dependence between events using mutually dependent point processes, with intensities depending on historical events.

\begin{figure}[!h]
\centering
\includegraphics[width=.8\linewidth]{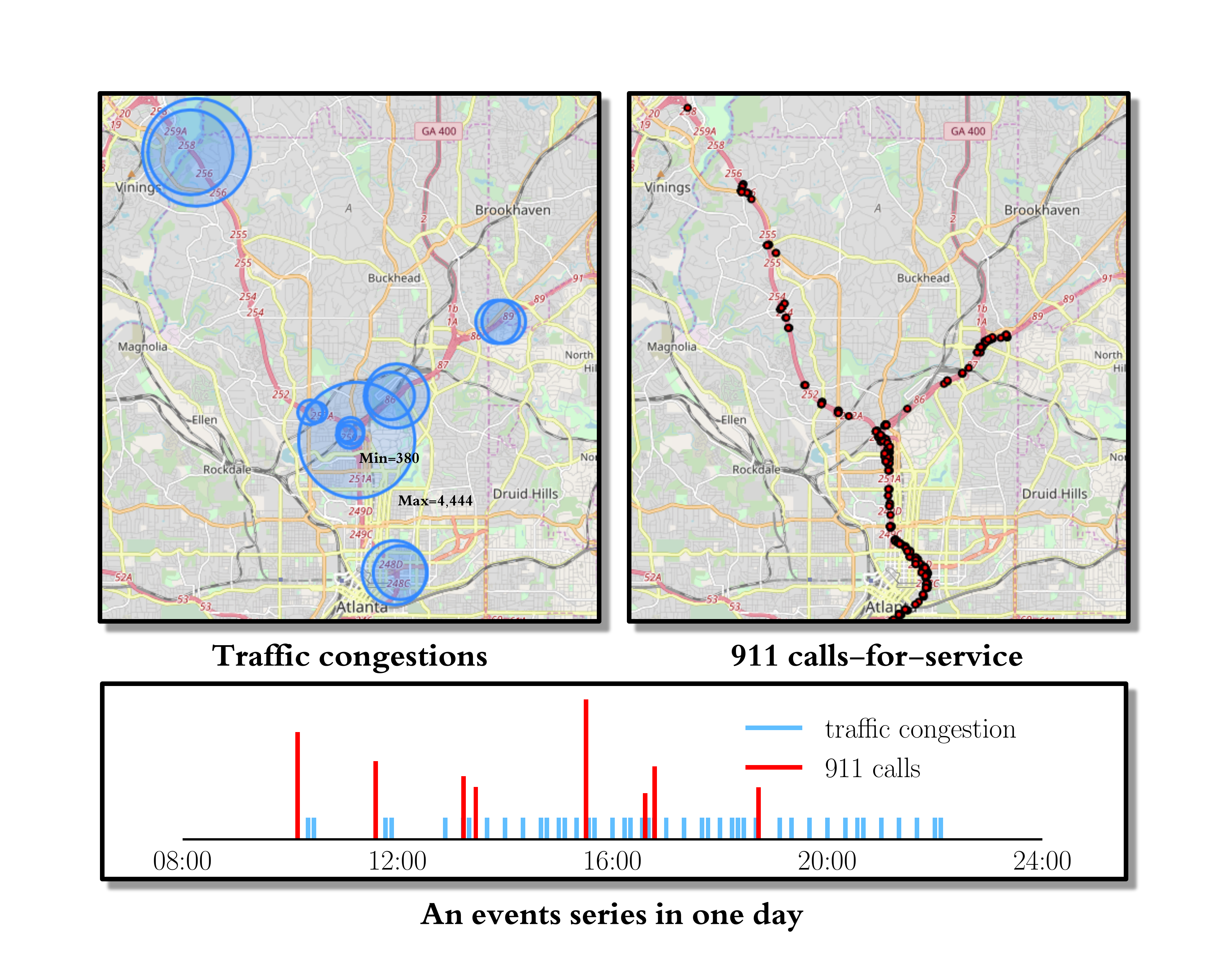}
\caption{An illustration of the Atlanta traffic dataset. The left panel shows the numbers of traffic congestions recorded by 14 traffic sensors. The size of the blue circle represents the number of traffic congestion events recorded by the traffic sensor. The right panel shows the spatial distribution of traffic incidents reported by 911 calls on the highway. Black dots represent the locations of traffic incidents. The panel in the bottom shows an event series in a single day. The height of the red bar indicates the length of the processing time.}
\label{fig:data-overview}
\vspace{-0.1in}
\end{figure}

There are two main reasons for the knowledge gap between existing point process models and our application in traffic congestion event modeling. (1) Most existing models assume that the influence function {\it decays monotonically} over time and space and introduce parametric models for the influence function. For instance, this approach is used in methods based on the popular Recurrent Neural Networks (RNNs) \cite{Du2016,Mei2017, Li2018, Upadhyay2018, Xiao2017A, Xiao2017B, Zhu2019B}, which can capture some complex temporal dependence with particular assumptions, e.g., \cite{Du2016} assumes that the influence of an event decreases exponentially over time. However, in traffic modeling settings, the influence of past events may not decay monotonically over time or space. For example, suppose that a major car accident occurs on the highway. The police are called to the scene, and they may need to wait for a specialized unit to move the wreckage. The process can take several hours during which the highway is blocked. In such a situation, the traffic event's influence may not decrease over a while and only start to relieve after the traffic blockage has been cleared. (2) We need to consider the special spatial correlation structure induced by road networks. Indeed, the spatial dependence is highly {\it directional} and what happens ``up-stream'' will influence what happens ``down-stream'', and the sensors along the same road (in the same direction) will have higher correlations. Existing Hawkes process models tend to discretize space and treat it as a multi-dimensional Hawkes process without considering such a special structure.

In this paper, we develop a novel spatio-temporal attention point processes (APP) model for traffic congestion events while capturing the relationship between exogenous and endogenous factors for traffic congestion. Specifically, we consider the police response to traffic incidents as \emph{exogenous promotion} \cite{rizoiu2017} on traffic congestion (since such events indicate traffic accidents) and consider the influence between congestion events as \emph{endogenous self-excitation}. To model the dynamics of the endogenous self-excitation, we borrow the idea of the so-called \emph{attention} mechanism \cite{Luong2015, Vaswani2017}, which is originally proposed to capture the non-linear dependence between words in natural language processing.  This approach allows us to go beyond the assumption that the influence of the historical event fades over time and leverage the attention mechanism to develop a flexible framework that ``focuses'' on past events with high importance scores on the current event. Moreover, we also introduce a new element in the attention mechanism, the adaptive \emph{score function} to measure the importance between past events and the current event, 
which is highly interpretable and adapts the conventional dot-product score \cite{Vaswani2017} to measure non-homogeneous spatio-temporal distance in traffic networks. To tackle the directional spatial correlation induced by road networks, we also adopt the idea of the ``tail-up'' model (developed for spatial statistics for Gaussian processes) to our point process setting. Finally, to achieve constant memory in the face of streaming data, we introduce an online algorithm to efficiently implement our APP model's attention component, where only the most informative events in the past are retained for computation. We show that our proposed method outperforms the state-of-the-art in achieving both a higher likelihood function of the point process model and higher prediction accuracy, using real-data traffic and police data set from Atlanta, USA.  
%
% \textcolor{red}{
The novelty of our paper can be highlighted as follows. Our APP model is the first attempt to combine traffic sensor count data with police reports to model traffic events, to the best of our knowledge. Our spatio-temporal model is tailored for our application by including a novel attention-based mechanism to capture the non-homogeneous spatio-temporal dependence of the historical events, and a ``tail-up'' model to capture the directional spatial dependence. 
% }
%The main contributions of our paper are as follows. (1) To the best of our knowledge, our APP model is the first attempt to combine traffic sensor count data with police reports for traffic event modeling. (2) In terms of methodology, our APP model includes a novel attention-based mechanism to capture a non-homogeneous spatio-temporal dependence of the event on the past. (3) The APP model includes a novel approach to capture the directional spatial dependence by adapting a similar idea used for the ``tail-up'' model, which was used to model spatial correlation for hydrology systems such as rivers and streams. (4) Our experimental results demonstrate the benefits of the APP model both on synthetic and real case studies.

%\vspace{-.05in}
%\noindent{\bf Related work.}

\subsection{Related work}

Earlier works \cite{Abadi2015, Lv2015, ma2017, cui2018, Liao2018, Yuan2018, Gu2019, Pan2019, Zheng2019, ZhuL2019} on traffic modeling focusing on predicting speed, volume, density, and travel time have achieved much success in this field. Other works \cite{wilson2001, zeroual2017, sole2016} targeting at modeling traffic congestion based on the speed and density of vehicle stream have resulted in good mathematical descriptions for traffic flow. However, {\it traffic events modeling} is much less studied and in the nascent stage.  Existing work in discrete event modeling using point processes have been used for modeling earthquake events, crime events, social networks, such as \cite{Hawkes1971, Gomez2010, Yuan2019, Zhu2019A}. Such works often make strong assumptions and specify a parametric form of the intensity functions. Although they enjoy good interpretability and are efficient, parametric models may not be expressive enough to capture the event dynamics in some applications. 

Recently there has been work on improving the expressive power of point process models. One approach is to develop RNNs based point process models \cite{Du2016, Mei2017, Xiao2017A, Zhu2019B}, which develop an RNN structure to mimic the conditional intensity function; however, the conditional intensity still discounts the influence of historical events. Another approach uses neural networks to directly model temporal dependence of sequence data. This includes \cite{Li2018, Xiao2017B}, which use RNNs as a generative model without specifying the conditional intensity function explicitly, and \cite{Omi2019}, which uses a neural network to parameterize the hazard function, and the conditional intensity can be derived by taking the derivative of the hazard function. However, the above approach focuses on temporal dependence rather than accounting for special structures in spatio-temporal dependence and is not directly applicable to our setting.

% There also have been some work that model stochastic processes using attention mechanisms \cite{Zhang2019, Kim2019}. \cite{Kim2019} uses self-attention to model a class of neural latent variable models, called Neural Processes \cite{Garnelo2018}, which is not for sequential data specifically.
A recent work \cite{Zhang2019} also uses attention to model the historical information in point processes. However, they still assume the conditional intensity function follows a parametric exponential decaying model, which may not capture distant events, although they can be important. We do not make such assumptions in our APP model and can capture important events as long as their ``importance score'' is high. Moreover, \cite{Zhang2019} focuses on temporal point processes, while we consider spatio-temporal point processes; they use the dot-product score function to measure the similarity of two events, while we develop a more flexible score function based on neural networks. 
Another related work \cite{Zheng2019} models traffic counts using two individual attention structures to embed spatial and temporal information; however, traffic count modeling is completely different from traffic event modeling considered in our paper.

\section{Datasets}
\label{sec:data}

In this section, we introduce the main datasets for our study, which consists of (1) a traffic congestion events dataset; (2) the 911 calls-for-service dataset; and (3) the Atlanta traffic network dataset. 

%In this section, we introduce a large-scale traffic dataset, which consists of three sub-datasets: (1) traffic congestion sub-dataset; (2) 911 call-for-service sub-dataset; and (3) traffic network sub-dataset. 

%In this section, we introduce unique large-scale traffic dataset,%\footnote{The dataset will be available after the review phase}, 
%which consists of three sub-datasets: (1) traffic congestion sub-dataset; (2) 911 call-for-service sub-dataset; and (3) traffic network sub-dataset. 

\subsection{Traffic congestion}
\label{sec:traffic-congestions}

The traffic congestion dataset is collected from the Georgia Department of Transportation (GDOT) \cite{GDOT}, which records the real-time traffic condition throughout the state of Georgia. These data are collected by traffic sensors installed on highways' main traffic points, which records the number of vehicles pass by the sensor every 5 minutes. The dataset also includes lane information (the number of lanes, the maximum speed of the lane, and the direction) at locations where the sensors are installed. The number of lanes can be used to estimate the maximum number of vehicles on the highway during that time. We assume the highway capacity --  the maximum number of vehicles that a highway can support in normal conditions -- is a linear function of the number of lanes. 

\begin{figure}[!t]
\centering
\begin{subfigure}[h]{0.49\linewidth}
\includegraphics[width=\linewidth]{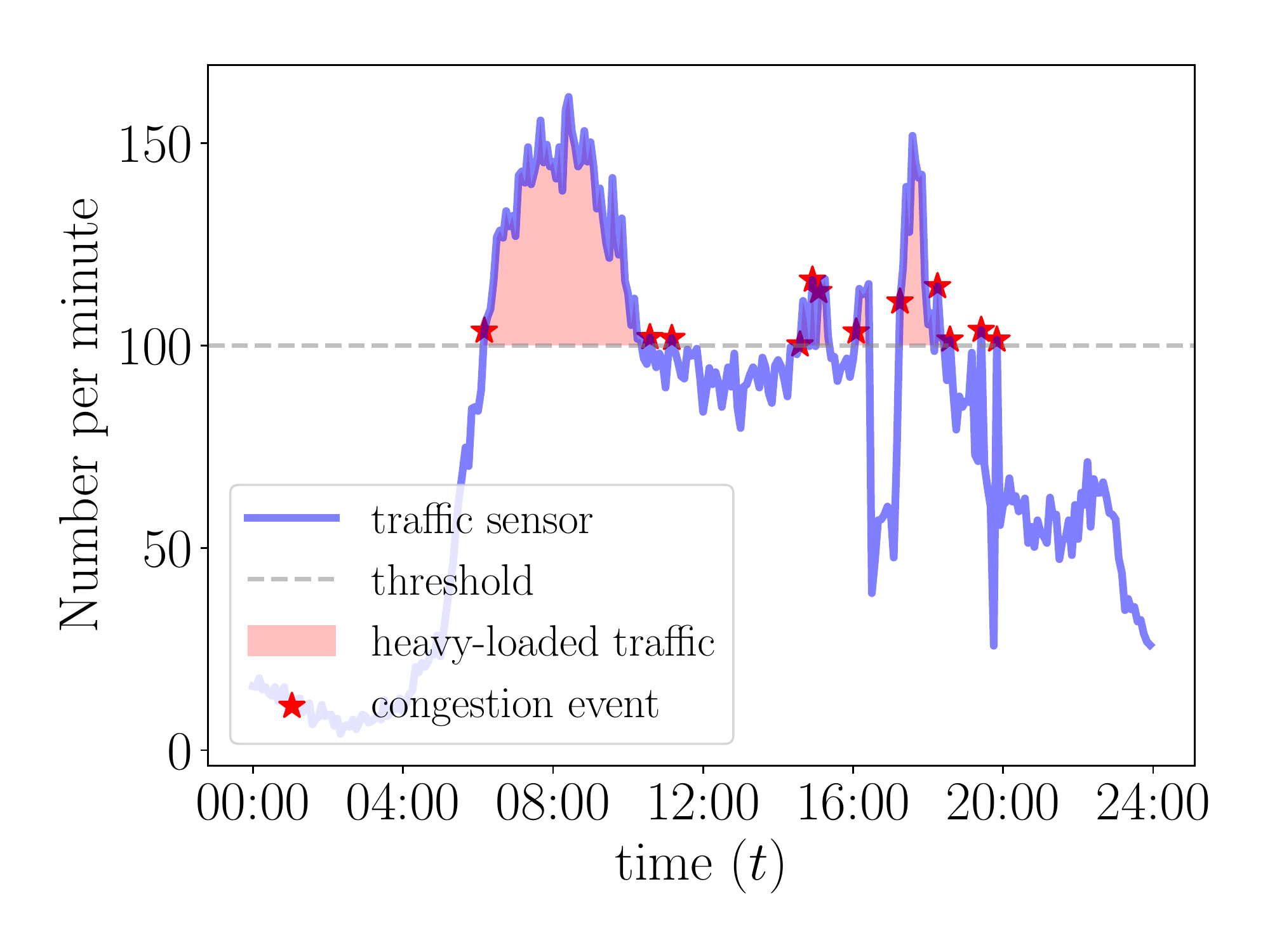}
\caption{Wednesday, April 4th, 2018}
\end{subfigure}
\begin{subfigure}[h]{0.49\linewidth}
\includegraphics[width=\linewidth]{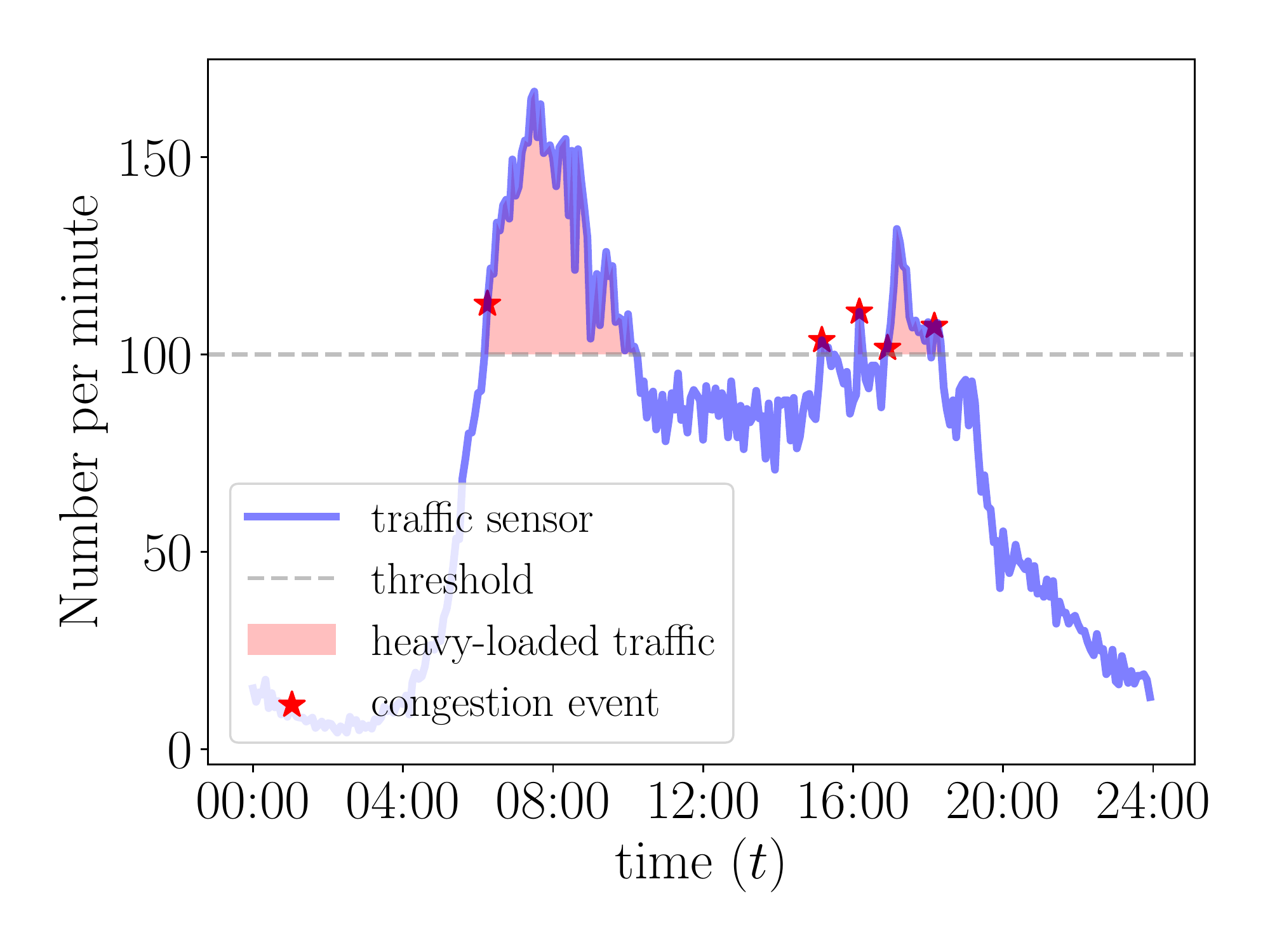}
\caption{Thursday, April 5th, 2018}
\end{subfigure}
\caption{Two examples of extracted traffic congestion recorded on the same traffic sensor labeled as ``\texttt{L1N}'' in Fig.~\ref{fig:traffic-network} for two consecutive days. The rise of the traffic counts above the normal traffic level is followed by a drop below the level.
}
\label{fig:traffic-events-exps}
\vspace{-0.1in}
\end{figure}

As recognized in the literature \cite{treiber2013traffic}, when traffic congestion initiates, the traffic count usually starts to increase and exceeds the normal volume. Then vehicles begin to slow down. As the congestion sets in, the throughput of the highway drops, and the traffic count falls below the normal level. We illustrate two examples in Fig.~\ref{fig:traffic-events-exps}, which are extracted from the Atlanta traffic data. We find such patterns prevalent in traffic congestion events -- the rise of the traffic counts above the normal traffic level is followed by a drop below the level. Thus, we detect a congestion event when the traffic counts exceed a threshold. Moreover,  we found the traffic congestion events tend to cluster in time -- in Fig.~\ref{fig:traffic-events-exps},  the red stars indicate when the traffic counts exceed the threshold and the ``stars'' are clustered in times; this is a motivation for us to consider the self-exciting process model in the paper.

% As commonly recognized in \cite{treiber2013traffic}, when the traffic congestion starts, it usually starts with a higher than usual volume. Then vehicles start to slow down, and congestion sets in, which causes the throughput of the highway drops as well, and the traffic count will fall below the threshold (the normal level). We illustrate this using two examples extracted from our real traffic counts data, as shown in Fig.~\ref{fig:traffic-events-exps}. The rising of the traffic counts above the threshold is followed by a lower than normal traffic level; these patterns together define a traffic congestion event. 
% Thus, we use the time when the traffic counts exceed the threshold to extract such type of ``event'' (red stars), which is often correlated with the onset of traffic congestion events. Here the red stars indicate when the traffic counts exceed the threshold (the normal traffic level): clearly, the ``stars'' are clustered in times; this is one of the motivations why we consider the self-exciting process model in the paper. 

\begin{figure}[!t]
\centering
\includegraphics[width=.95\linewidth]{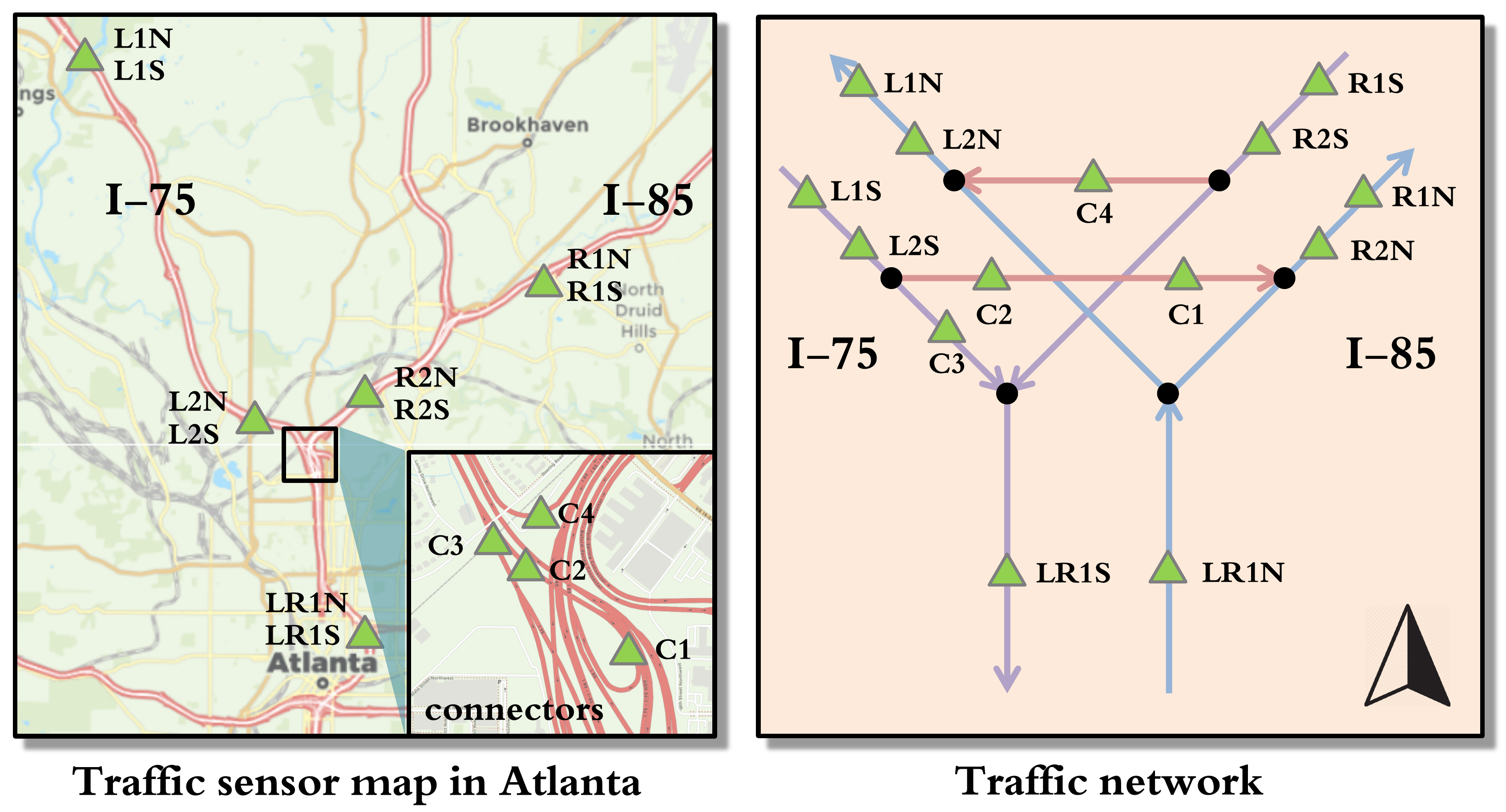}
\caption{The traffic network for major highways in Atlanta. The left panel shows the spatial distribution of traffic sensors, where green triangles represent locations of traffic sensors. Traffic sensors on the highway are bi-directional, i.e., two directions of the same location have separate traffic sensors to monitor the traffic condition. The right panel shows the traffic network and where traffic sensors located on the network. Each line segment represents one specific road segment, and black dots represent the confluence of two roads.}
\label{fig:traffic-network}
\vspace{-0.1in}
\end{figure}

For our study, we consider $K=14$  traffic sensors on two major highways in metro Atlanta, I-75 and I-85, shown in Fig.~\ref{fig:traffic-network}. We denote their geo-locations (latitude and longitude) on the traffic network by $r_k \in \mathscr{S} \subset \mathbb{R}^2$, $k=1, \ldots, K$, where $\mathscr{S}$ denote the location space of traffic networks, which will be discussed in Section~\ref{sec:traffic-network}. 
Let $\{x_i\}_{i=1}^{N_x}$ represent a sequence of traffic congestion events in a single day, where $N_x$ is the number of the congestion events generated in one-day $[0, T)$. The $i$-th congestion event $x_i = (t_i, s_i)$ is a data tuple consisting of the occurrence time $t_i \in [0, T)$, and the sensor location index $s_i \in \{1, \ldots, K\}$. We extract 18,618 traffic congestion events from 174 days between April 2018 and December 2018. The maximum and the minimum number of events in a single day are 168 and 19, respectively.

\subsection{911 calls-for-service}
\label{sec:911-calls}

As mentioned in Section~\ref{sec:intro}, traffic incidents may trigger unexpected congestion over traffic networks. Since the police respond to many traffic incidents, the 911 calls-to-service contains useful traffic incidents records. Thus, we use another data set, the 911 calls-for-service reports provided by the Atlanta Police Department (also used in \cite{Zhu2019A, Zhu2019C}) and extract all traffic incidents. When a 911 call reporting a traffic incident is received, the operator will start a record ID and dispatch police offices to the scene. The record includes the \emph{call time} and occurrence location.  The police officers arrive at the scene and start the investigation. Once the police complete the investigation and clean the scene, a \emph{clear time} will be added to the record. The time interval between the call time and the clear time is called \emph{processing time}. A long processing time is usually a strong indicator of a severe traffic incident that significantly impacts the highway traffic condition.

Let $\{y_j\}_{j=1}^{N_y}$ represent a sequence of traffic incidents extracted from 911 calls in a single day, where $N_y$ is the total number of recorded 911-call incidents in one day. The $j$-th 911-call incident $y_j = (t_j, r_j, z_j)$ is a data tuple consisting of the call time $t_j \in [0, T)$, the occurrence location $r_j \in \mathscr{S}$ on the traffic network, and the processing time $z_j >0$ indicating the police processing time. We extract 19,805 such 911-call records on two major highways between April 2018 and December 2018 with processing time larger than 15 minutes.  The incidents include many different categories ranging from speeding tickets to massive car pile-ups.

\subsection{Traffic network}
\label{sec:traffic-network}

Due to the nature of the traffic flow, there is a strong spatial dependence among the traffic data collected at different locations on the network. Moreover, the network topology and the direction of the flow will impact spatial correlations. For instance, there should not be a correlation for data collected at two locations that do not share a common traffic flow. In metro Atlanta, there are two major highways I-75 and I-85, passing through the city center. These two highways start from the northwest and northeast side of the city, run due south, and intersect in Midtown, as shown in the left of Fig.~\ref{fig:traffic-network}. Between I-75 and I-85, two connectors bridge two highways via single-direction ramps. 
We extracted the network information of I-75, I-85 and their connectors in Atlanta from OpenStreetMap \cite{OpenStreetMap}, which is an editable map database and allows us to construct, visualize, and analyze complex traffic networks. The traffic network of a city is represented by a set of road segments defined in the OpenStreetMap dataset as shown in the right of Fig.~\ref{fig:traffic-network}. Let $\mathscr{S} \subset \mathbb{R}^2$ represent the set of all geo-locations on the network. Suppose there are $L$ road segments, and the set of locations on each segment is denoted as $\mathscr{S}_l \subset \{1, \ldots, L\}$, $l = 1, \ldots, L$. For any location $s\in\mathscr{S}$ on the network, we define the upstream portion $\vee_s \subseteq \{1, \ldots, L\}$ of the network to include $s$ itself and all locations upstream from $s$. Define the downstream portion $\wedge_s \subseteq \mathscr{S}$ to include $s$ itself and all locations downstream from $s$. For two locations $u, v \in \mathscr{S}$, the distance $d(u,v) \in \mathbb{R}^+$ is defined as the stream distance along the highway if one of the two locations belongs to the downstream of the other.  We also denote $u \rightarrow v$ when $v$ belongs to $\vee_{u}$ and the two points are said to be \emph{flow-connected}. When two points are \emph{flow-unconnected}, neither $u$ belongs to $\wedge_v$ nor $v$ belongs to $\wedge_u$, and the relationship between $u$ and $v$ is denoted $u \centernot \rightarrow v$.

% \vspace{-0.1in}
\section{Methodology}
\label{sec:methods}
%\vspace{-0.1in}

\begin{figure}[!b]
\vspace{-0.1in}
\centering
\includegraphics[width=.8\linewidth]{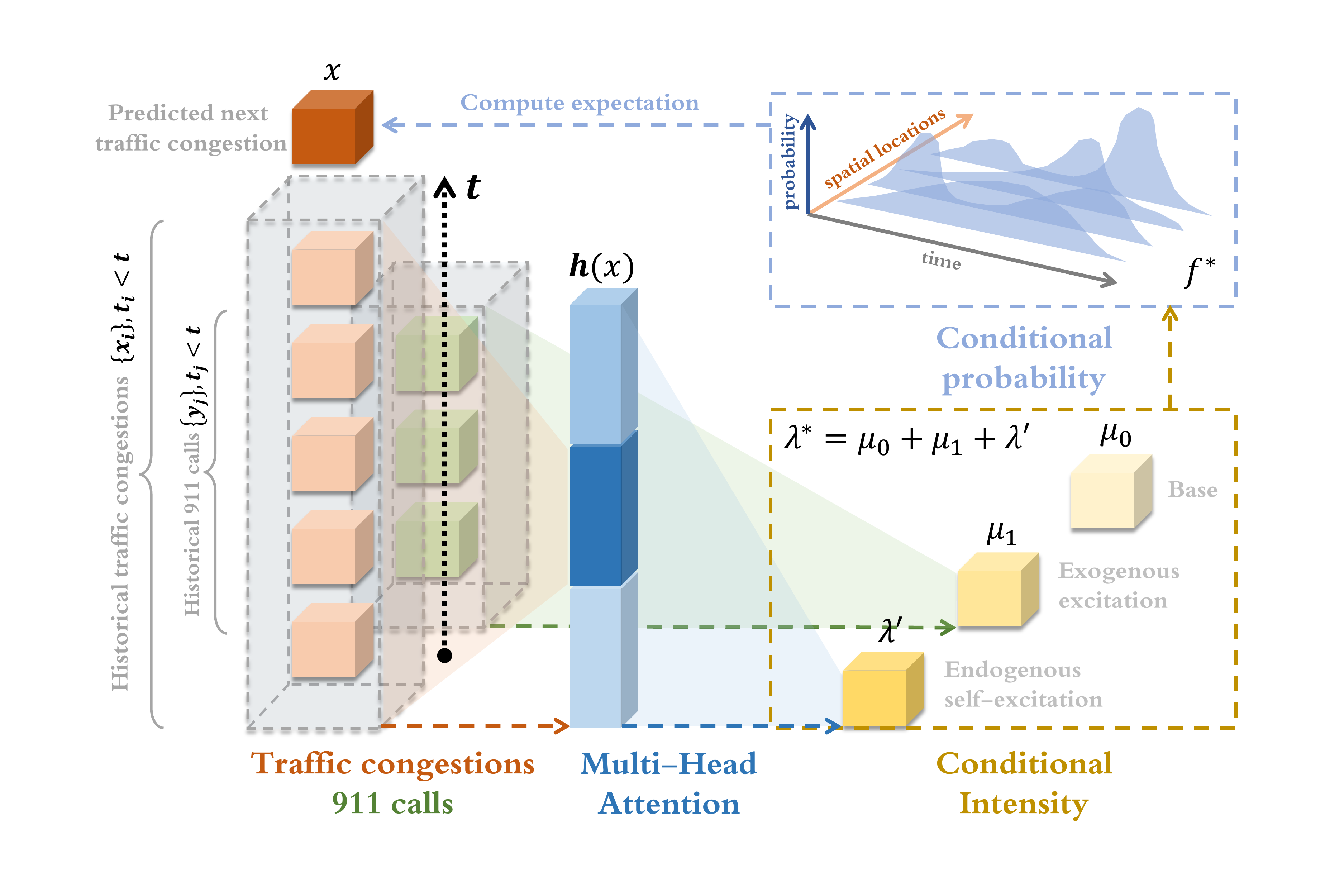}
\caption{The architecture for traffic congestion events modeling. The historical data, consisting of traffic congestion events and the 911 incidents, is fed into the multi-head attention mechanism to estimate the conditional intensity function for the point process model. The conditional intensity consists of three key components: (1) The background intensity $\mu_0$ (which is estimated from the traffic data); (2) The exogenous promotion $\mu_1$ (representing the influence of traffic events from police reports); (3) Endogenous self-excitation $\lambda'$ (capturing the spatio-temporal dependence between congestions events).}
\label{fig:architecture}
\end{figure}

In this section, we propose an attention-based point process model for modeling traffic congestion events while consider the police 911-call data. The architecture of our model is shown in Fig.~\ref{fig:architecture}.

\subsection{Spatio-temporal point processes}
\label{sec:point-processes}

The spatio-temporal point processes (STPPs) \cite{Reinhart2017} consist of a sequence of events, represented by their times, locations, and marks. In our setting, the events are traffic congestion events. Let $\{x_1 , x_2 , \dots, x_{N_x}\}$ represent a sequence of congestion events sampled from a STPP and $\{y_1, y_2, \dots, y_{N_y}\}$ represent a sequence of 911-call incidents recorded in the same time period. Recall that $N_x$ and $N_y$ are the number of congestion events (extracted from traffic counts) and the number of traffic incidents (extracted from 911-calls), respectively, in the time horizon $[0, T)$. Let $\mathcal{H}_t = \mathscr{X}_t \cup \mathscr{Y}_t$ denote the $\sigma$-algebra generated by all historical events before time $t$, where $\mathscr{X}_t = \{x_i\}_{t_i<t}$ and $\mathscr{Y}_t = \{y_j\}_{t_j<t}$.
The distribution in STPPs are characterized via a conditional intensity function $\lambda(t, k|\mathcal{H}_t)$, which is the conditional probability of observing an event at $(t,k) \in [0, T) \times \{1,\dots,K\}$, given the events' history $\mathcal{H}_t$. Formally, 
$
%\begin{equation*}
    \lambda(t, k | \mathcal{H}_t)
    % = \frac{\mathbb{E}\left[ N_{k}([t, t+dt)| \mathscr{H}_t \right]}{dt},
    = \mathbb{E}\left[ N_{k}([t, t+dt)) | \mathcal{H}_t \right]/dt,
   % \label{eq:def-conditional-intensity}
%\end{equation*}
$
where $N_{k}(A)$ is the counting measure of events for sensor $k$ over any set $A \subseteq [0, T)$. 

To capture the dynamic between traffic congestion and traffic incidents, we consider the following form for the conditional intensity function in our model:
\begin{equation}
    \lambda(t, k|\mathcal{H}_t) = 
    \underbrace{\mystrut{2.ex} \mu_0(t, k)}_{\substack{\text{background}\\\text{intensity}}} + 
    \underbrace{\mystrut{2.ex} \mu_1(t, k|\mathscr{Y}_t)}_{\substack{\text{exogenous}\\\text{promotion}}} + 
    \underbrace{\mystrut{2.ex} \lambda'(t, k|\mathscr{X}_t)}_{\substack{\text{endogenous}\\\text{self-excitation}}}, 
    \label{eq:lambda}
\end{equation}
where $\mu_0(t, k)$ is the background intensity at $(t, k)$, which can be estimated from data. The exogenous promotion $\mu_1(t, k|\mathscr{Y}_t)$ captures the influence of past 911-call incidents reported by the police before time $t$. The endogenous self-excitation $\lambda'(t, k|\mathscr{X}_t)$ captures the influence of past traffic congestion before time $t$. In the remainder of the section, we will discuss how to model these two types of triggering effects in Section~\ref{sec:intervention} and Section~\ref{sec:attention-pp}, respectively.

% \vspace{-0.1in}
\subsection{Police intervention as exogenous promotion}
\label{sec:intervention}

As we discussed in Section~\ref{sec:data}, police response to traffic incidents is often correlated with an increase in the strain on urban traffic. Such strain only spreads along the traffic direction and decays as the distance increases from where the traffic incident originated. Due to this consideration, the spatial correlation between two locations $u, v \in \mathscr{S}$ on the highway is determined by the traffic network structure and their ``stream distance'' $d(u, v)$, which is the distance to reach from $u$ to $v$ in the directed graph. Also, the spatial correlation may vary from time to time since the traffic intensity is always changing throughout the day. Here, we denote such spatial correlation between two arbitrary locations $u, v\in \mathscr{S}$ on the traffic network at time $t \in [0,T)$ as $\alpha_t(u, v) \in [0, 1]$. We will discuss the estimation of the spatial correlation in Section~\ref{sec:tail-up}.

Now we consider the traffic incidents extracted from the police reports, $y_j = (t_j, r_j, z_j)$, at $(t, k)$ as an additive exogenous promotion when $t$ is in the middle of process of the event $y_j$, i.e., $t \in [t_j, t_j + z_j)$. Formally, the exogenous promotion in \eqref{eq:lambda} can be defined as
\begin{equation}
\mu_1(t, k | \mathscr{Y}_t) = \sum_{y_j \in \mathscr{Y}_t} \gamma \alpha_t(r_k, r_j) \cdot \delta_t\big([t_j, t_j + z_j)\big)   
\end{equation}
where $\delta_t(A)$ is the Dirac measure, i.e., taking the value 1 when $t \in A$ and 0 otherwise. The parameter $\gamma >0$ captures the decay rate of the influence.  

\subsection{Attention-based self-excitation modeling}
\label{sec:attention-pp}

The idea of Attention Point Processes (APP) is to model the non-linear and long-term dependence of the current traffic congestion event from past congestion events using the attention mechanism \cite{Luong2015, Vaswani2017}. The attention model has recently become popular in natural language processing and computer vision to tackle non-linear and long-range dependencies. Still, it has been rarely used outside of these domains. We adopt the attention mechanism for point process models by redefining ``attention'' in our context and adopt the ``multi-heads'' mechanism \cite{Vaswani2017} to represent more complex non-linear triggering effects. Specifically, we model the endogenous self-excitation $\lambda'(t, k|\mathscr{X}_t)$ in \eqref{eq:lambda} using the output of the attention structure. The exact definition of the conditional intensity is as follows.

%The idea of Attention-based Point Processes (APP) is to model the nonlinear dependence of the current traffic congestion event from past congestion events using the attention mechanism \cite{Luong2015, Vaswani2017}. Specifically, we model the endogenous self-excitation $\lambda'(t, k|\mathscr{X}_t)$ in \eqref{eq:lambda} using the output of the attention structure. 
% We also consider multi-heads in the attention mechanism \cite{Vaswani2017}, which offers multiple ``representation subspace'' for events in the sequence. 
%We also found it beneficial to consider multi-heads in the attention mechanism \cite{Vaswani2017}, which linearly project the data tuple of events with different, learned linear projections to $M$ ``representation subspace'', respectively. On each of these projected representations of events, we then perform the attention function in parallel, yielding output values. These outputs are concatenated and once again projected, resulting in the final values. Multi-head attention allows the model to jointly attend to information from different representation subspaces at different positions and enjoys greater expressive power in representing the non-linear triggering effects. The exact calculation of the conditional intensity is carried out as follows.

\begin{figure}[!t]
\centering
\includegraphics[width=.7\linewidth]{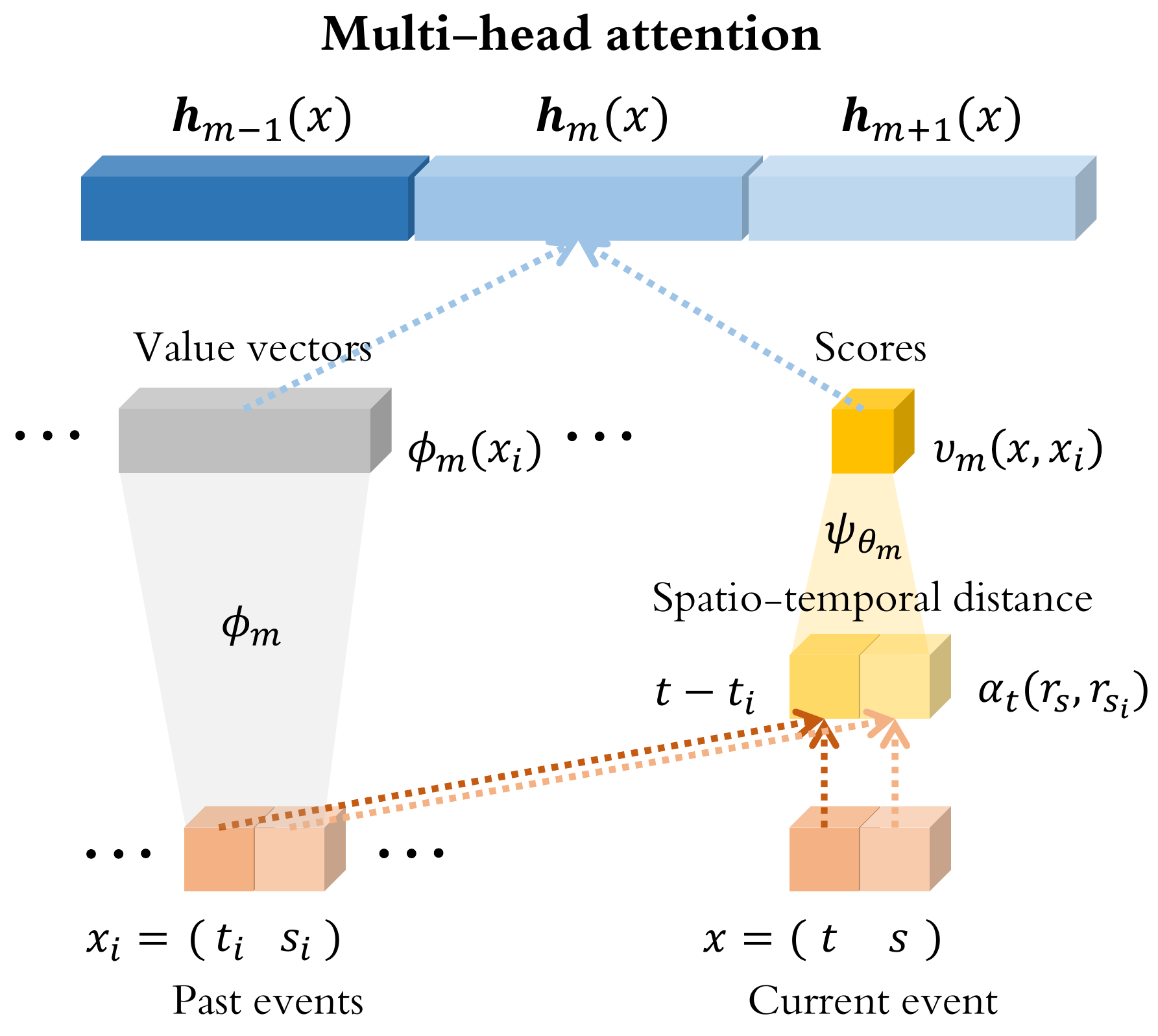}
\caption{The multi-head attention architecture.}
\label{fig:multi-head-attention}
\vspace{-0.1in}
\end{figure}

For the notational simplicity, let $x \coloneqq (t, s)$ represent the data tuple of the current congestion event and $x' \coloneqq (t', s') \in \mathscr{X}_{t}, t' < t$ represent the data tuple of another event in the past. 
As shown in Fig.~\ref{fig:multi-head-attention}, for the $m$-th attention head, we score the current congestion event against its past event, denoted as $\upsilon_m(x, x') \in \mathbb{R}^+$. For the event $x$, the score $\upsilon_m(x, x')$ determines how much \emph{attention} to place on the past event $x'$ as we encode the history information, which will be further discussed in Section~\ref{sec:kernelized-score}. 
The normalized score $\widetilde\upsilon_m(x, x') \in [0, 1]$ for events $x$ and $x'$ is obtained by employing the softmax function as:
\begin{equation}
  \widetilde\upsilon_m(x, x') = \frac{
  \upsilon_m(x, x')}{
  \sum_{x_i \in \mathscr{X}_{t}} \upsilon_m(x, x_i)},
  \label{eq:score}
\end{equation}   

The $m$-th {\it attention head} $\boldsymbol{h}_m(x) \in \mathbb{R}^p$ for the event $x$ is defined as % multiplying each value embedding by the score and adding them up, i.e.,
\begin{equation}
    \boldsymbol{h}_m(x) = \sum_{x_i \in \mathscr{X}_{t}} \widetilde\upsilon_m(x, x_i) \phi_m(x_i),
    \label{eq:attention}
\end{equation}   
where the value embedding $\phi_m(x) = x^\intercal W_m^v$ and $W_m^v \in \mathbb{R}^{d \times p}$ is a weight matrix, $d$ is the data dimension (here $d=3$), and $p$ is the dimension of value embedding.
In comparison with the \emph{self-attention} mechanism \cite{Vaswani2017}, the current event $x$ and past event $x_i$ are analogous to \emph{query} and \emph{key}, respectively, and the value embedding for the past event $\phi_m(x_i)$ is analogous to \emph{value}. 
The multi-head attention $\boldsymbol{h}(x) \in \mathbb{R}^{Mp}$ for event $x$ is the concatenation of $M$ single attention heads:
\[
    \boldsymbol{h}(x) = \text{concat}\left(\boldsymbol{h}_1(x), \dots, \boldsymbol{h}_M(x)\right).
\]

Denote $\lambda'(t, k|\mathscr{X}_t)$ in \eqref{eq:lambda} as $\lambda'(x|\mathscr{X}_t)$. The historical information in $\mathscr{X}_t$ can be modeled as a non-linear transformation of the multi-head attention $\boldsymbol{h}(x)$. Hence, the endogenous self-excitation $\lambda'$ can be specified as:
\begin{equation}
\begin{aligned}
    \lambda'(x | \mathscr{X}_t) 
    & = \lambda'(x | \boldsymbol{h}(x)) = \text{softplus}\left(\boldsymbol{h}(x)^\intercal W + b\right),
    \label{eq:lambda-prime}
\end{aligned}
\end{equation}
where $W \in \mathbb{R}^{Lp}$ is the weight matrix and $b \in \mathbb{R}$ is the bias term. The function $\text{softplus}(x) = \log(1 + e^x) > 0$ is a smooth approximation of the ReLU function, which ensures that the intensity strictly is positive at all times. 
 
\subsection{Score function}
\label{sec:kernelized-score} 

The score function quantifies how likely a past event triggers one event.  The most commonly used score function in attention models is the dot-product score function. 
Typically, the score is obtained by computing the inner product between the query $x$ and the key $x'$ given two mapping matrices $W, W'$ (the so-called key/query embedding), i.e., $\upsilon(x, x') = x^\intercal W^\intercal W' x'$. This can be viewed as their inner product in the embedding space.
However, as discussed in Section~\ref{sec:intervention}, for our setting, the correlation between two locations on the highway may not depend on their Euclidean distance, and this correlation may also vary over time. We adopt the spatial correlation $\alpha_t(r_{s_n}, r_{s_i})$ at time $t$ between locations of two events $r_{s_n}, r_{s_i} \in \mathscr{S}$ rather than their Euclidean distance. 
% The estimation of the spatial correlation is discussed in Section~\ref{sec:tail-up}. 

\begin{figure}[!b]
\vspace{-0.1in}
\centering
\includegraphics[width=.8\linewidth]{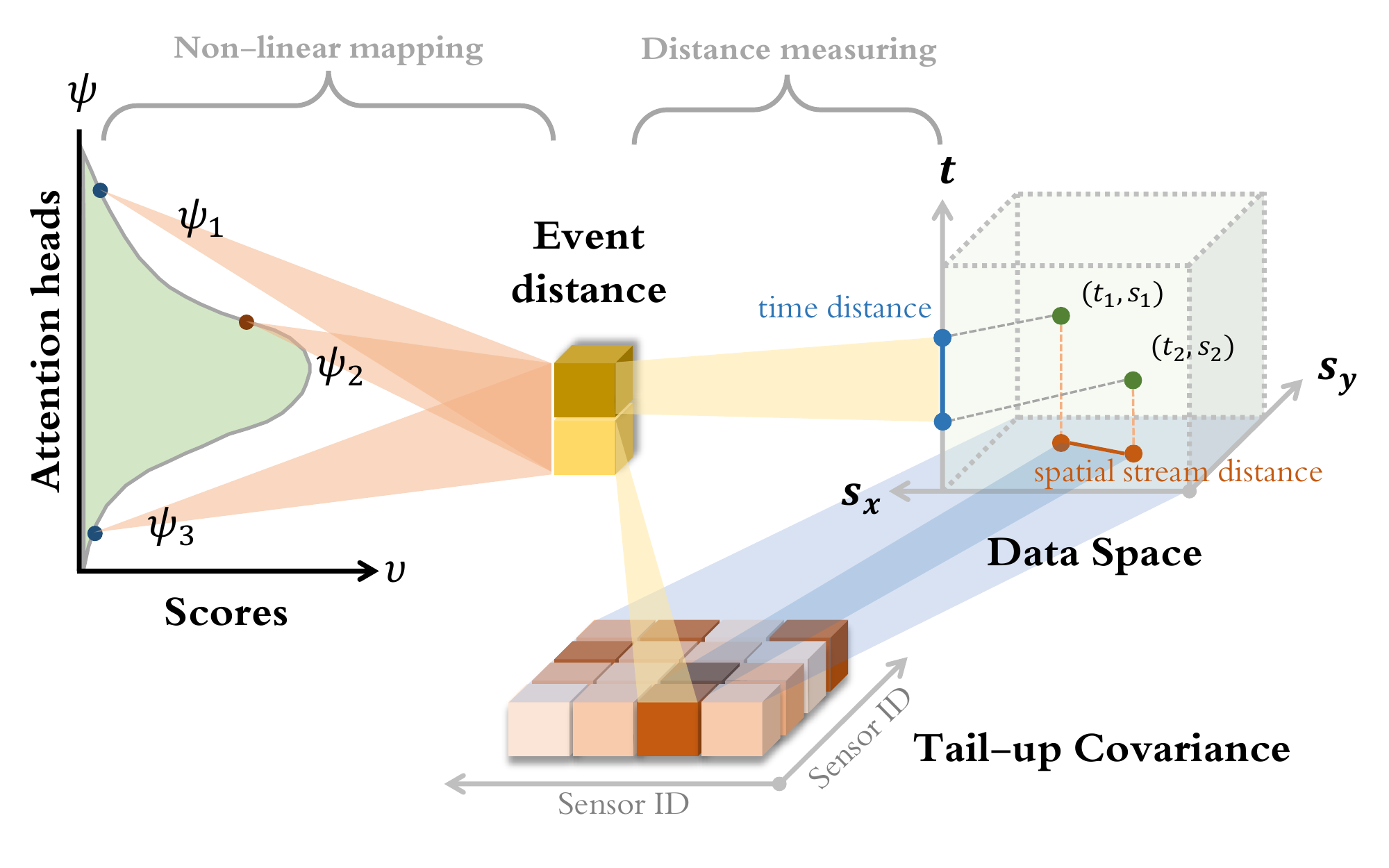}
\caption{The illustration of the score calculation. First, we measure the distance between two events, in which their spatial correlation is represented by our tail-up model; Then we calculate the score by transforming their spatio-temporal distance via a non-linear mapping. Note that we use different non-linear mapping for different attention head.}
\label{fig:score}
% \vspace{-0.1in}
\end{figure}

As illustrated in Fig.~\ref{fig:score}, the score function $\upsilon_m(x, x')$ for the $m$-th attention head can be expressed as:
\begin{equation}
    \upsilon_m(x, x') = 
    \psi_{\theta_m} \bigg(t - t', \alpha_{t}(r_{s}, r_{s'})\bigg),
    \label{eq:score-detail}
\end{equation}
where the $\psi_{\theta_m}(\cdot, \cdot) \in \mathbb{R}^+$ is a multi-layer neural network parameterized by a set of weights and biases denoted by $\theta_m$. The neural network takes the spatial correlation $\alpha_{t}(r_{s}, r_{s'})$ and the temporal distance $t - t'$ as input and yields a non-negative score. The score function can be interpreted as a weighted spatio-temporal distance, where the weights of time and space are learned from data. Note that the score function for each attention head may be different.

%\vspace{-0.1in}
\subsection{Tail-up spatial model in score function} 
\label{sec:tail-up}

To capture the spatial correlation $\alpha_t(u,v)$ between two locations $u, v \in \mathscr{S}$ on the traffic network at time $t \in [0, T)$, we adopt the idea of tail-up spatial model, which is originally based on moving averages \cite{Barry1996, Garreta2010, Hoef2006, Hoef2010} and widely used in river systems modeling \cite{Chen2017, Cressie2006, Peterson2007}. For the notational simplicity, we omit $t$ in the following. There are three major advantages of tail-up models against other methods: (1) The tail-up models use stream distance rather than the Euclidean distance, which is defined as the shortest distance along with the traffic network between two locations. (2) The statistical independence is imposed on the observations located on stream segments that do not share the same traffic flow. (3) Proper weighting is incorporated on the covariance matrix entries when the road segments in the network are splitting into multiple segments to ensure that the resulting covariance is stationary. The tail-up models are appropriate when we want to enforce zero covariance when locations are flow-unconnected. This situation can occur when the flow dominates a variable (e.g., when traffic congestion enters a stream and can influence the upstream, it induces correlation in sensor measurements only when they are flow-connected).

Assume the traffic at location $u \in \mathscr{S}$ can be viewed as a white-noise random process $Z_u$, where observable locations on the traffic network can be developed by creating random variables as the integration of a moving average function over the process on the upstream of the network \cite{Cressie2006},
\[
  Z_u = \mu_u + \int_{\vee_u} g(r - u) \sqrt{\frac{w(r)}{w(u)}} dB(r),
\]
where $\mu_u$ is a deterministic mean process at location $u$, $\vee_u$ denotes the upstream of location $u$ and $w(r) = w^l$ for all location $r\in\mathscr{S}_l$ on the segment $l$, which is the weight that satisfies the additivity constraint to ensure the stationarity in variance. The weights can be estimated using normalized average traffic counts for each segment on the traffic network. The moving average function $g(\cdot)$ is square-integrable and defined on $\mathbb{R}$. The $B(r), r\in \mathscr{S}$ is a Brownian process starting from sources of the traffic network, progressing toward the outlets,  separating or merging themselves at traffic forks. The spatial correlation $\alpha(u,v)$ is obtained by $\text{cov}(Z_u, Z_v) = \mathbb{E}[Z_u Z_v] - \mathbb{E}[Z_u] \mathbb{E}[Z_v]$, i.e.,
\[
  \alpha(u, v) = \int_{\vee_u \cap \vee_v} g(r - u) g(r - v) \frac{w(r)}{\sqrt{w(u)w(v)}} d r.
\]
Let $\Delta r$ be a stream distance on $\mathbb{R}^+$ and define $C(\Delta r) \coloneqq \int_{\mathbb{R}} g(r) g(r - \Delta r) dr$. By choosing a particular moving average function, we can reparameterize $C(\cdot)$ in forms typically seen in the spatial statistical literature. We adopt the tail-up exponential model here \cite{Hoef2010}, i.e.,
\[
  C(\Delta r) \coloneqq \beta \exp(- \Delta r / \sigma),
\]
where $\beta, \sigma$ are parameters of the tail-up model. 
Let $d(u, v) \in \mathbb{R}^+$ be the stream distance between locations $u, v \in \mathscr{S}$ on the traffic network.
The above covariance can be simplified as:
\begin{equation}
  \alpha(u, v) = 
  \begin{cases}
    C\left( d(u, v) \right)\sqrt{\frac{w(u)}{w(v)}} , & u \rightarrow v\\
    0, & u \centernot \rightarrow v.
  \end{cases}
  \label{eq:alpha}
\end{equation}
Note that the tail-up model simplifies the parameterization of spatial correlation in the score function by replacing two complete weight matrices to be only with two trainable parameters $\beta, \sigma$.

\begin{figure}[!t]
\centering
\begin{subfigure}[h]{0.46\linewidth}
\includegraphics[width=\linewidth]{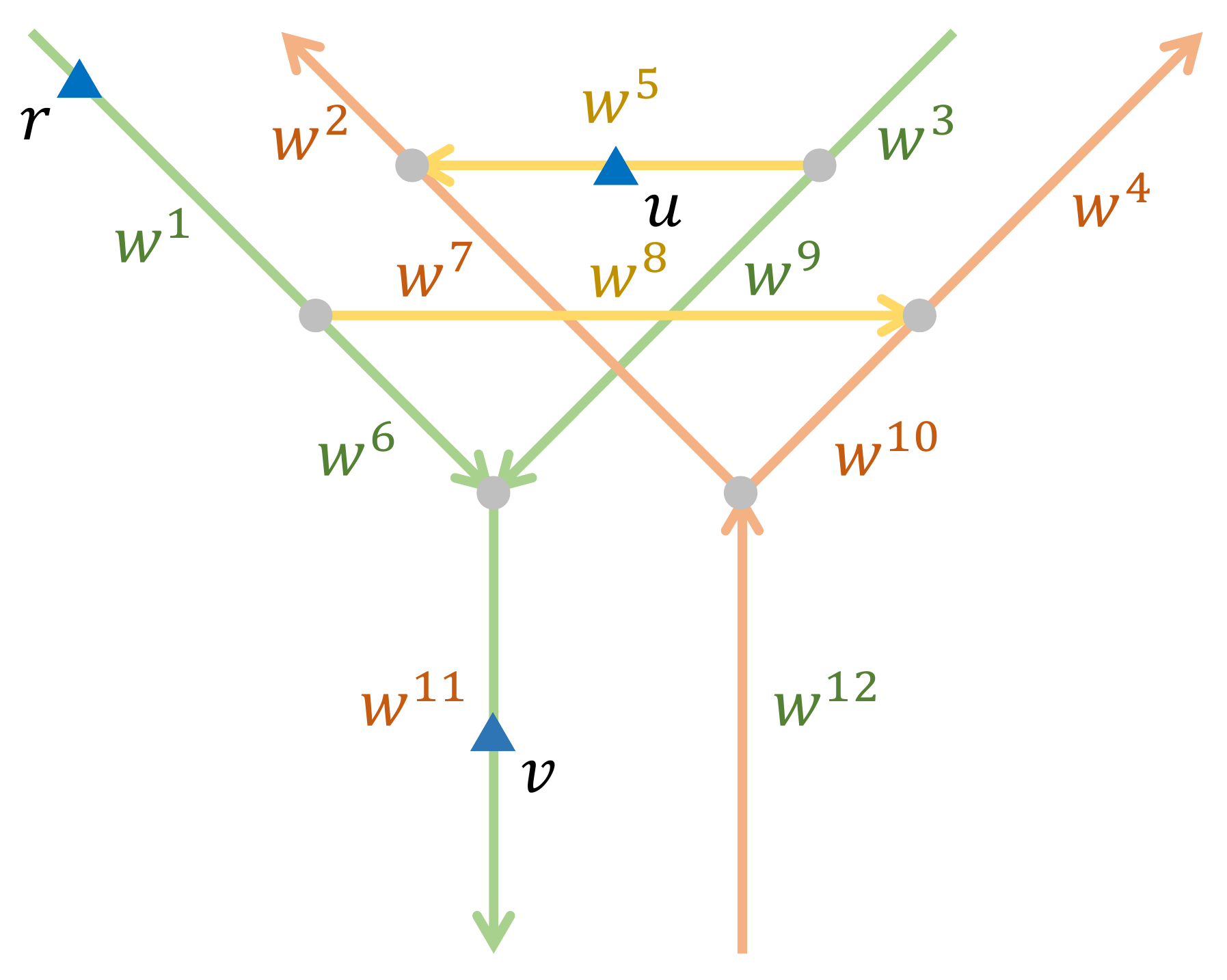}
\caption{Weight example}
% \label{fig:gan-macys}
\end{subfigure}
\begin{subfigure}[h]{0.51\linewidth}
\includegraphics[width=\linewidth]{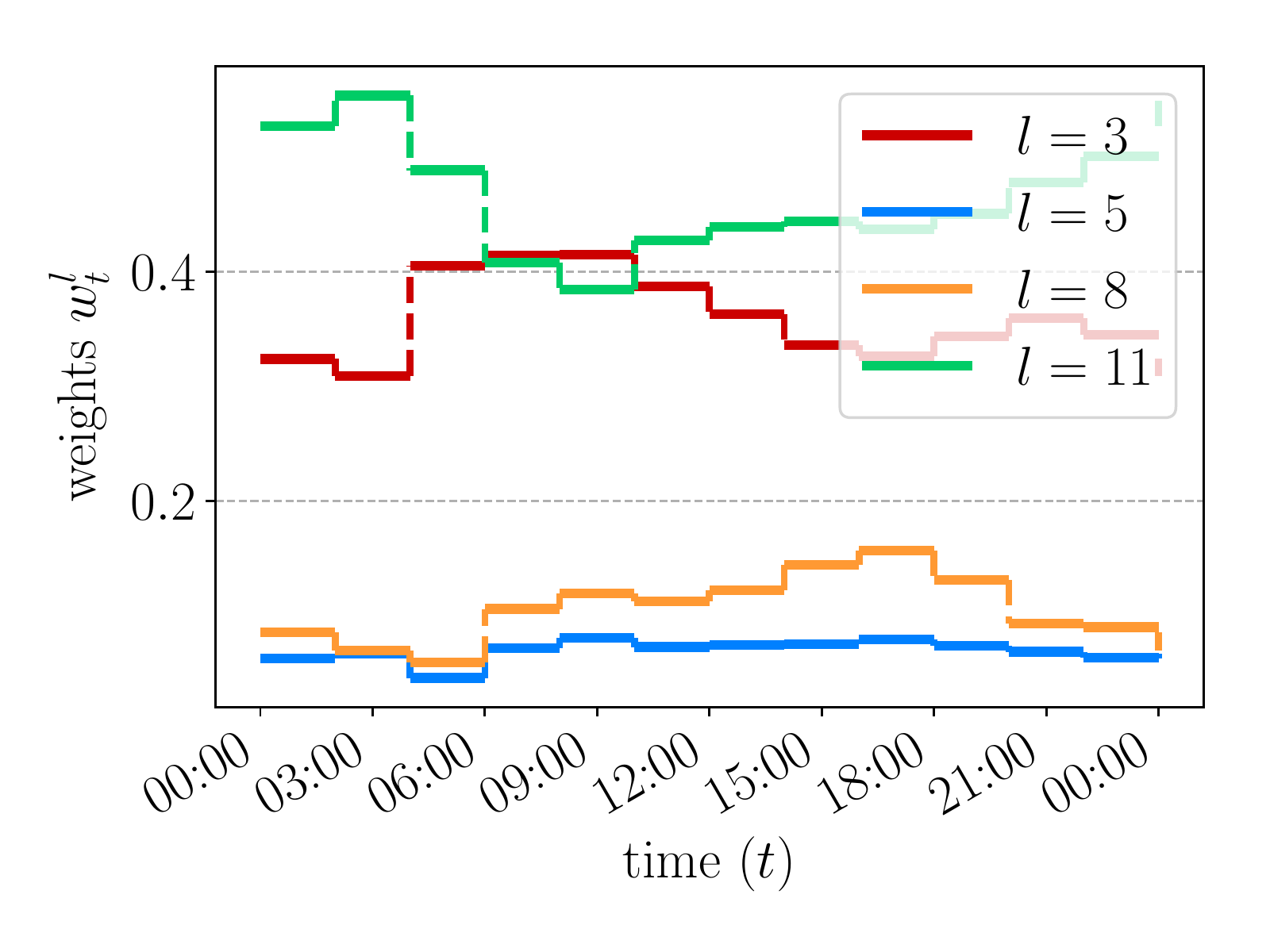}
\caption{Estimated weights $w_t^l$}
% \label{fig:gan-earthquake}
\end{subfigure}
\caption{Weighting example for the tail-up model: arrows indicate flow directions of segments on the traffic network; gray dots represent confluences; $w^l$ is the weight of the segment $l$; triangle $u$, $v$, and $r$ are observable locations.}
\label{fig:tail-up-exp}
\vspace{-0.1in}
\end{figure}

For example in Fig.~\ref{fig:tail-up-exp}, we first define all weights for each segment, i.e., $\{w^l\}_{l=1}^L$. 
The structure of the traffic network ensures that the sum of weights that flow into a confluence equals to the sum of weights that flow out of the same confluence (e.g. $w^1 = w^6 + w^7$ and $w^6 + w^9 = w^{11}$). Then we can obtain 
\[\alpha(u, r) = C(d(u, v)) \sqrt{w(u) / w(v)} = C(d(u, v)) \sqrt{w^5 / w^{11}},\] and $\alpha(u, r) = 0$ because locations $u$ and $r$ are flow-unconnected.

In the experiment, we consider the tail-up model to be time-variant, i.e., for each time $t$, we can obtain a set of weights $\{w_t^l\}_{l\in\mathscr{L}}$ and spatial correlation $\alpha_t(u, v)$ by estimating the traffic counts at time $t$, since the distribution of the traffic flow may vary over time. For example, as shown in Fig.~\ref{fig:tail-up-exp} (b), the time-variant weights of four traffic sensors are estimated based on the normalized traffic counts for every two hours. 

% \vspace{-0.1in}
\subsection{Online attention for streaming data}
\label{sec:online-attention}

For streaming data, calculating attention may be computationally intractable since the past events may accumulate as time goes on. Here, we propose an adaptive online attention algorithm to address this issue. The main idea is only to keep a fixed number of ``important'' historical events with high average scores when calculating the attention head. Using both synthetic and real data, we will show that, in many cases, only a few events have a dominant influence on future events. The online APP model can be performed efficiently using a small part of historical data.

The procedure for collecting ``important'' events in each attention head is as follows. When the $i$-th event occurs, for a past event $x_j, t_j < t_i$ in the $m$-th attention head, we denote the set of its score against the future congestion events $\{\widetilde\upsilon_m(x_i, x_j)\}_{i:t_j \le t_i}$ as $\mathscr{C}_{m,j}$. Then the average score of the event $x_j$ can be computed by $\bar\upsilon_{m,j} = (\sum_{\upsilon \in \mathscr{C}_{m,j}} \upsilon)/|\mathscr{C}_{m,j}|$, where $|\mathscr C|$ denotes the number of elements in a set $\mathscr C$. Hence, a recursive definition of the set $\mathscr{A}_{m,n}$ for selected events in the $m$-th attention head up until the occurrence of the event $x_i$ is written as: 
\begin{align*}
    & \mathscr{A}_{m,i} = \mathscr{X}_{t_{i+1}}, & \forall i \le \eta, \\
    & \mathscr{A}_{m,i} = \mathscr{A}_{m,i-1} \cup \{x_i\} \setminus \underset{j: t_j < t_i}{\arg \min} \left\{ \bar\upsilon_{m,j} \right\}, & \forall i > \eta,
\end{align*}
where $\eta$ is the maximum number of events we will retain. The complete event selection procedure is explained  Algorithm~\ref{alg:online-attention}. 
To perform the online attention, we substitute $\mathcal{H}_{t}$ in \eqref{eq:score} and \eqref{eq:attention} with $\mathscr{A}_{m,i}$ for all attention heads, where $i$ is the number of occurred events before time $t$.

\begin{algorithm}[!t]
    \caption{Event Selection for Online Attention}
    \label{alg:online-attention}
 \begin{algorithmic}
     \STATE{{\bfseries Input:} data $\boldsymbol{x} = \{x_{i}\}_{i \ge 1}$ and threshold $\eta$.}
    %  \STATE{Initialize $i=0, \mathscr{A}_{i,l} = \emptyset$.}
     \STATE{Initialize $\mathscr{A}_{m,0} = \emptyset$.}
     \FOR{$i=1$ {\bfseries to} $+\infty$.}
     \FOR{$m=1$ {\bfseries to} $M$.}
     \STATE{$\mathscr{A}_{m,i} \leftarrow \mathscr{A}_{m,i-1} \cup \{x_i\}$.}
    %  \STATE{// update the average score for each event in the past.}
     \STATE{Initialize $\mathscr{C}_{m,i} = \emptyset$.}
     \FOR{$j=1$ {\bfseries to} $i-1$}
     \STATE{$\mathscr{C}_{m,j} \leftarrow \mathscr{C}_{m,j} \cup \widetilde\upsilon_m(x_i, x_j)$.}
     \STATE{$\bar\upsilon_{m,j} \leftarrow (\sum_{\upsilon \in \mathscr{C}_{m,j}} \upsilon) / |\mathscr{C}_{m,j}|$.}
     \ENDFOR
    %  \STATE{// update the set after the $\eta$-th event.}
     \IF{$i > \eta$}
     \STATE{$\mathscr{A}_{m,i} \leftarrow \mathscr{A}_{m,i-1} \setminus \underset{j: t_j < t_i}{\arg \min} \left\{ \bar\upsilon_{m,j} \right\}$.}
     \ENDIF
     \ENDFOR
     \ENDFOR
\end{algorithmic}
\end{algorithm}

% \vspace{-0.1in}
\subsection{Learning and Inference}
\label{sec:learning}

Note that our model is jointly parameterized by a set of parameters $\{W, b, \gamma, \beta, \sigma, \{\theta_m, W_m^v\}_{m=1}^M\}$. We fit the model by the standard maximizing log-likelihood approach, which can be solved conveniently via the stochastic gradient descent. Equipped with the definition of conditional intensity in \eqref{eq:lambda}, we can write down the likelihood function of the data explicitly. Suppose there are $n$ observed events before $T$, denoted as $\boldsymbol{x} = \{(t_i, s_i)\}_{i=1}^{n}$. Let $F^*(t, k) = \mathbb{P}\{t > t_i, k | \mathcal{H}_{t}\}$ be the conditional cumulative function at $(t, k)$, where $t_i$ is the occurrence time of the last event before time $t$. Let $f^*(t, k)$ be the corresponding conditional density probability. For simplicity, denote the conditional intensity function $\lambda(t, k | \mathcal{H}_t)$ as $\lambda^*(t, k)$. The conditional intensity function for an arbitrary sensor $k$ is defined by $\lambda^*(t, k) = f^*(t, k) / (1 - F^*(t, k))$. From the definition above, we can show $\lambda^*(t, k) = - d \log(1 - F^*(t, k)) / dt$ and hence, $\int_{t_{n}}^t \lambda^*(\tau, k) d\tau = -\log(1 - F^*(t, k))$, where $F^*(t, k) = 0$, since the $(n+1)$-th event does not exist at time $t_{n}$. Therefore, $F^*(t, k) = 1 - \exp\{-\int_{t_{n}}^t \lambda^*(\tau, k) d\tau\}$ and
\begin{equation}
\begin{aligned}
  f^*(t, k) = \lambda^*(t, k) \cdot \exp\{-\int_{t_{n}}^t \lambda^*(\tau, k)d\tau\}.
  \label{eq:def-conditional-prob}
\end{aligned}
\end{equation}
Then the log-likelihood of observing a sequence $\boldsymbol{x}$ can be written as:
\begin{equation}
\begin{aligned}
  \ell(\boldsymbol{x}) 
  & = \sum_{i=1}^{n} \log \lambda^*(t_i, s_i) - \sum_{k=1}^K \int_0^T \lambda^*(t, k) dt.
  \label{eq:log-likelihood}
\end{aligned}
\end{equation}

Note that we can further predict a future event $(\hat{t}_{n+1}, \hat{s}_{n+1})$  given the past observations $\{x_i\}_{i=1,\dots,n}$ by using the conditional probability defined in \eqref{eq:def-conditional-prob} as follows:
\begin{equation}
    \begin{bmatrix}
        \hat{t}_{n+1} \\
        \hat{s}_{n+1}
    \end{bmatrix} = 
    \begin{bmatrix}
        \int_{t_{n}}^{T} \tau \sum_{k=1}^K f^*(\tau, k)d\tau\\
        \argmax_{k} \int_{t_{n}}^{T} f^*(\tau, k) d\tau
    \end{bmatrix}.
    \label{eq:prediction}
\end{equation}
In general, the integration above cannot be obtained analytically. Therefore, we use standard numerical integration techniques to calculate the expectation.

\begin{figure*}[!t]
\centering
\begin{subfigure}[h]{0.22\linewidth}
\includegraphics[width=\linewidth]{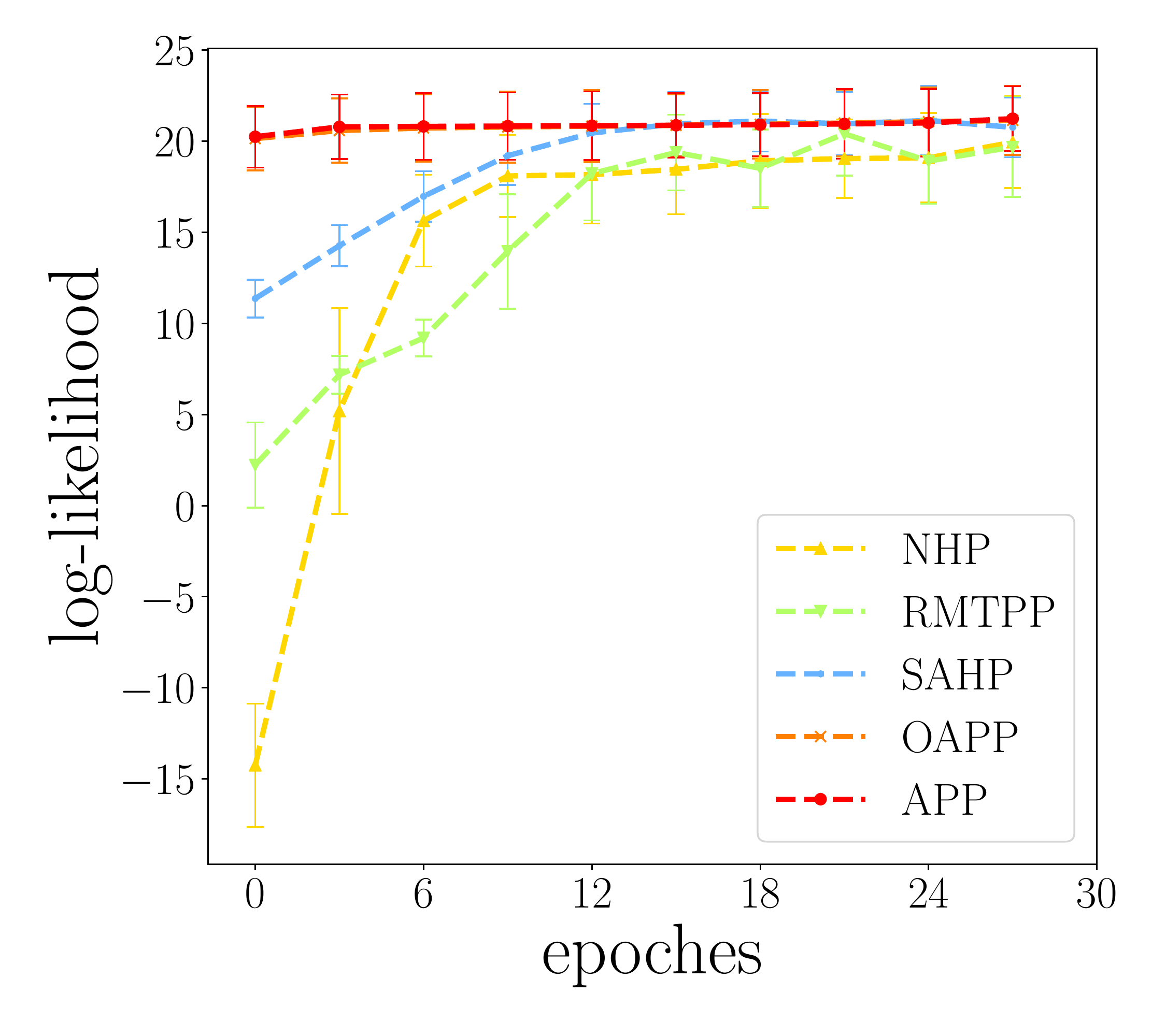}
\caption{Hawkes}
% \label{fig:gan-macys}
\end{subfigure}
% \hfill
\begin{subfigure}[h]{0.22\linewidth}
\includegraphics[width=\linewidth]{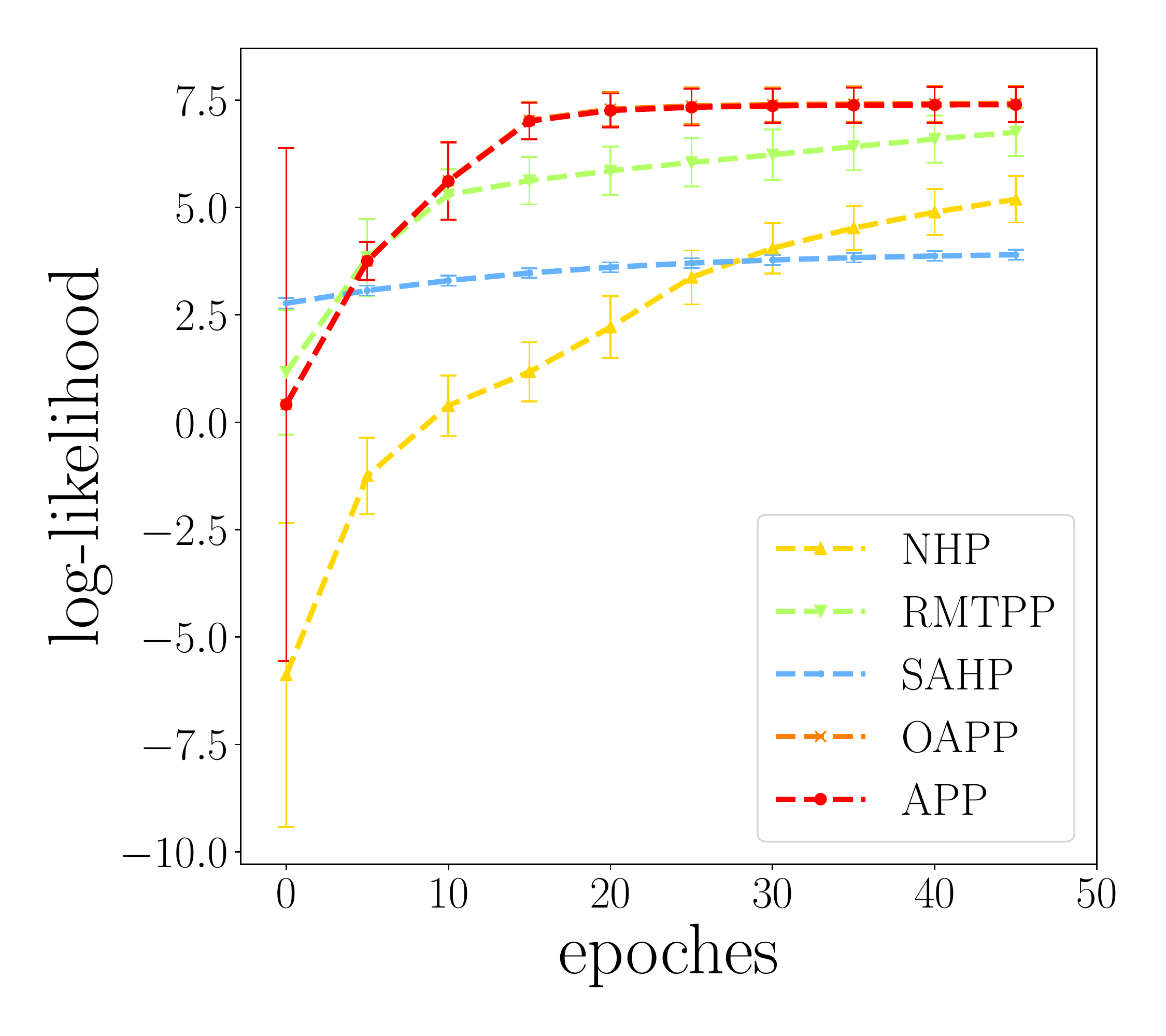}
\caption{self-correcting}
% \label{fig:gan-earthquake}
\end{subfigure}
% \hfill
\begin{subfigure}[h]{0.22\linewidth}
\includegraphics[width=\linewidth]{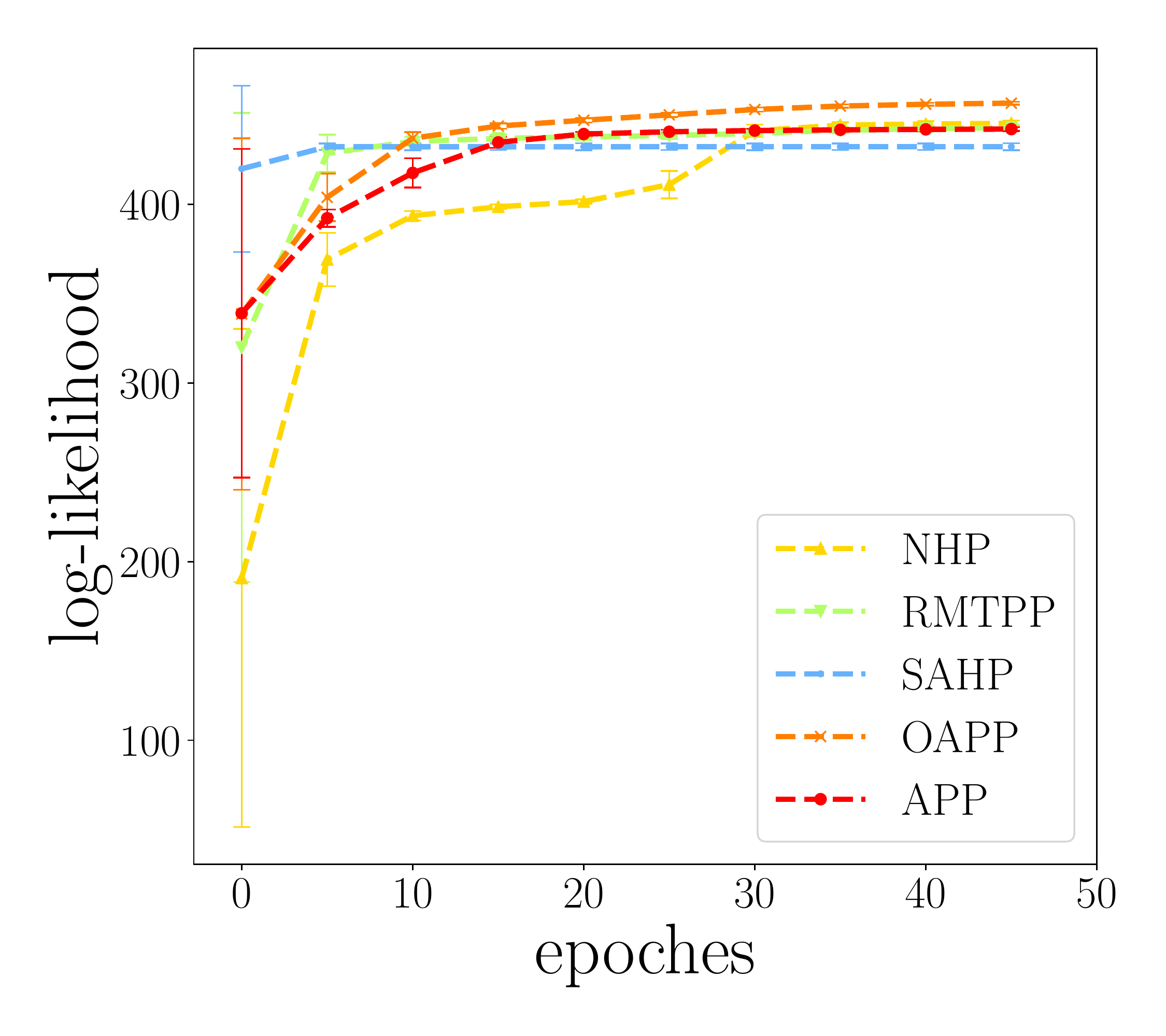}
\caption{non-homogeneous}
% \label{fig:gan-macys}
\end{subfigure}
% \hfill
\begin{subfigure}[h]{0.22\linewidth}
\includegraphics[width=\linewidth]{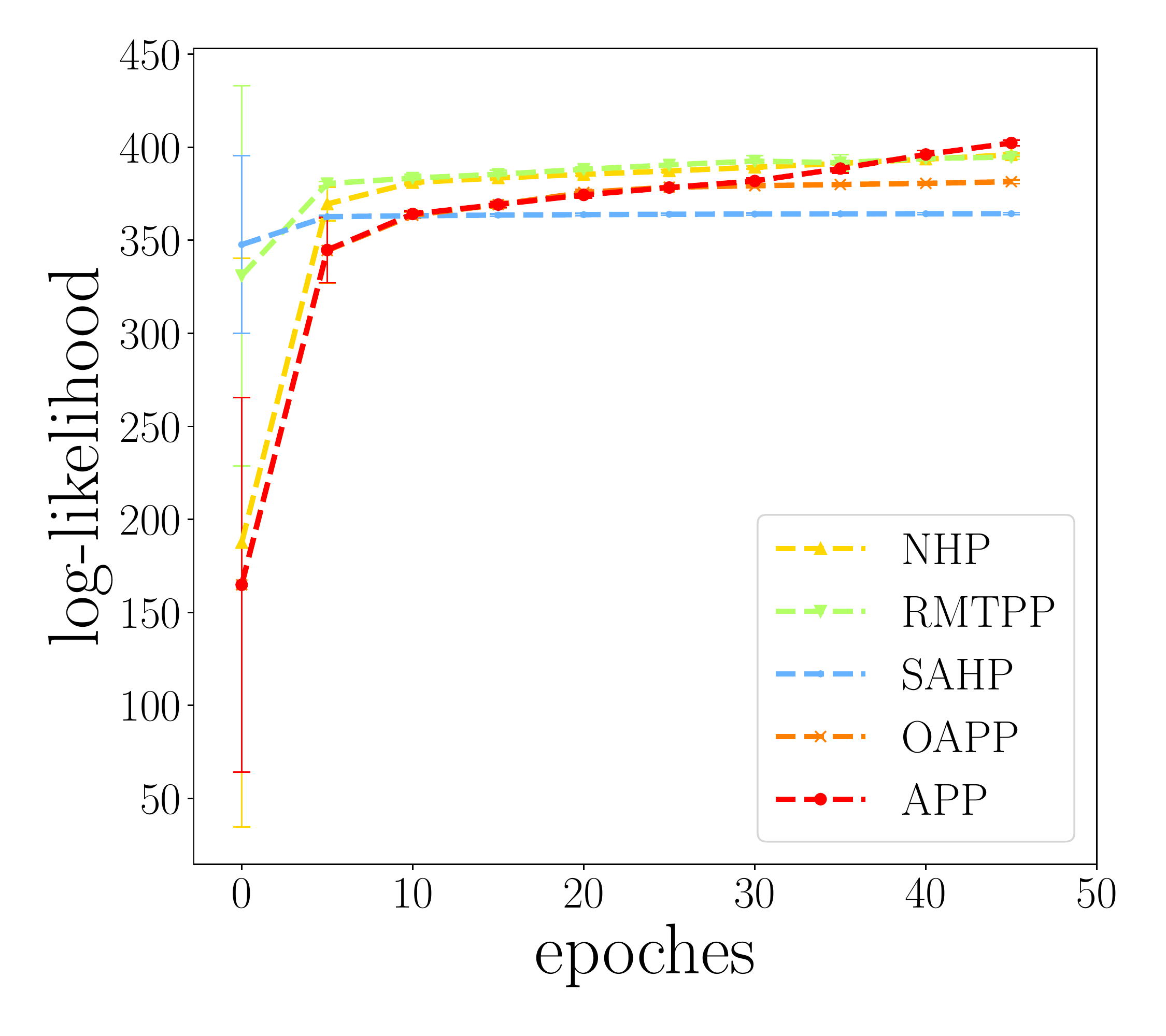}
\caption{non-homogeneous}
% \label{fig:gan-earthquake}
\end{subfigure}
\caption{The average log-likelihood of synthetic datasets versus training epochs. For each synthetic dataset, we evaluate the performance of the five methods based on the maximum log-likelihood averaged per series calculated for the test data.}
\label{fig:sim-res-loglik}
\vspace{-0.2in}
\end{figure*}

\begin{figure*}[!t]
\centering
\begin{subfigure}[h]{0.22\linewidth}
\includegraphics[width=\linewidth]{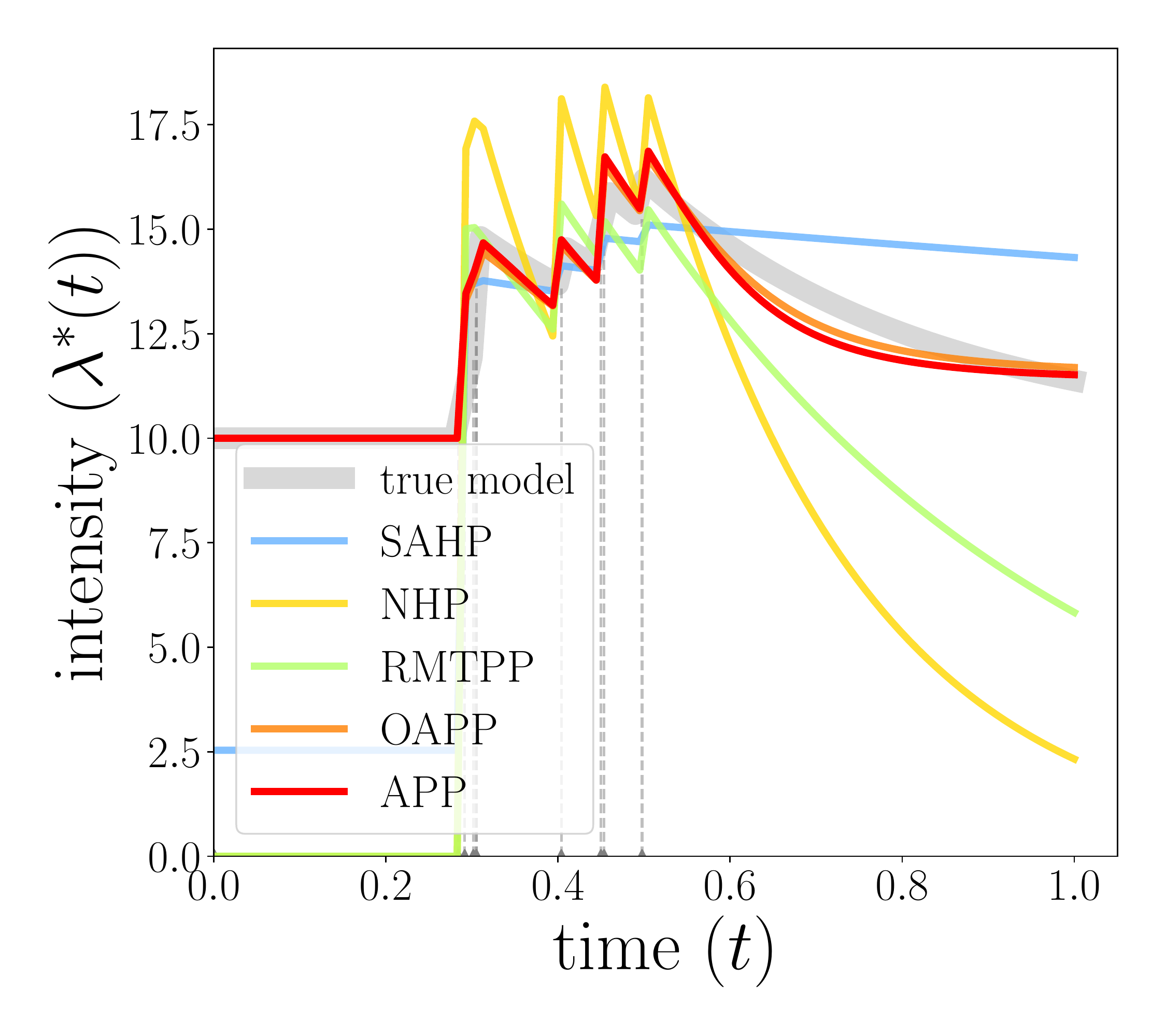}
\caption{Hawkes}
% \label{fig:gan-macys}
\end{subfigure}
% \hfill
\begin{subfigure}[h]{0.22\linewidth}
\includegraphics[width=\linewidth]{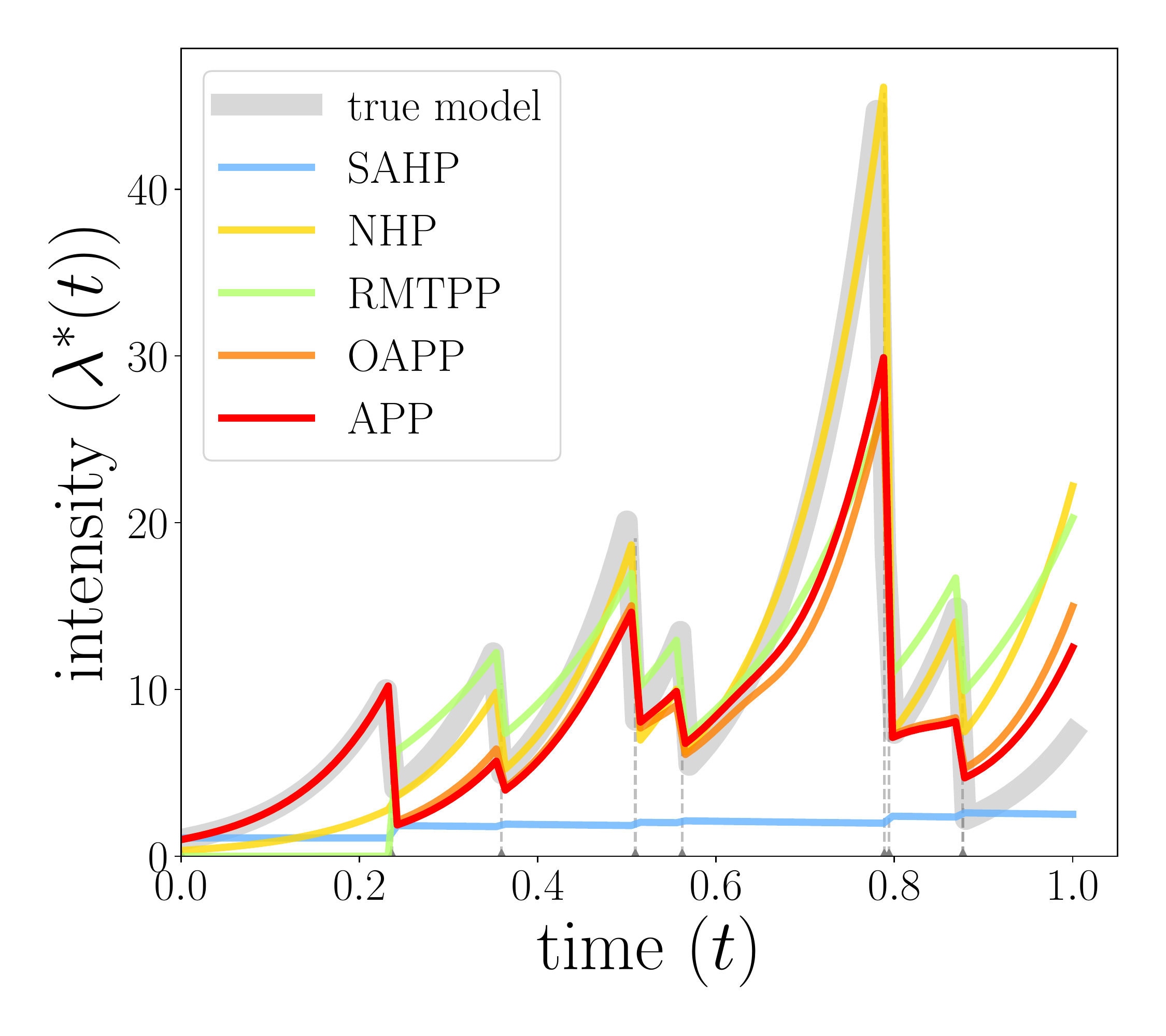}
\caption{self-correcting}
% \label{fig:gan-earthquake}
\end{subfigure}
% \hfill
\begin{subfigure}[h]{0.22\linewidth}
\includegraphics[width=\linewidth]{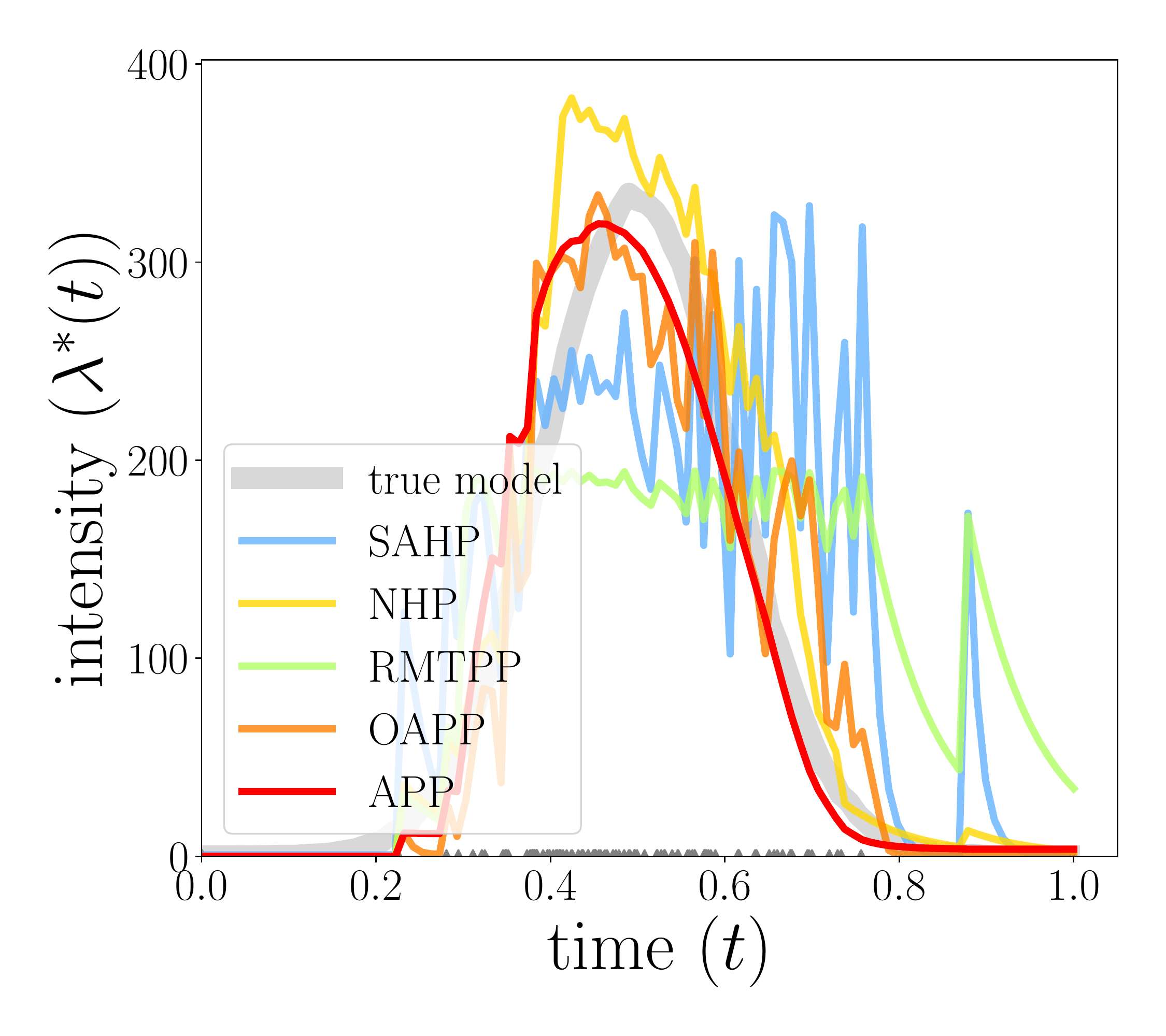}
\caption{non-homogeneous}
% \label{fig:gan-macys}
\end{subfigure}
% \hfill
\begin{subfigure}[h]{0.22\linewidth}
\includegraphics[width=\linewidth]{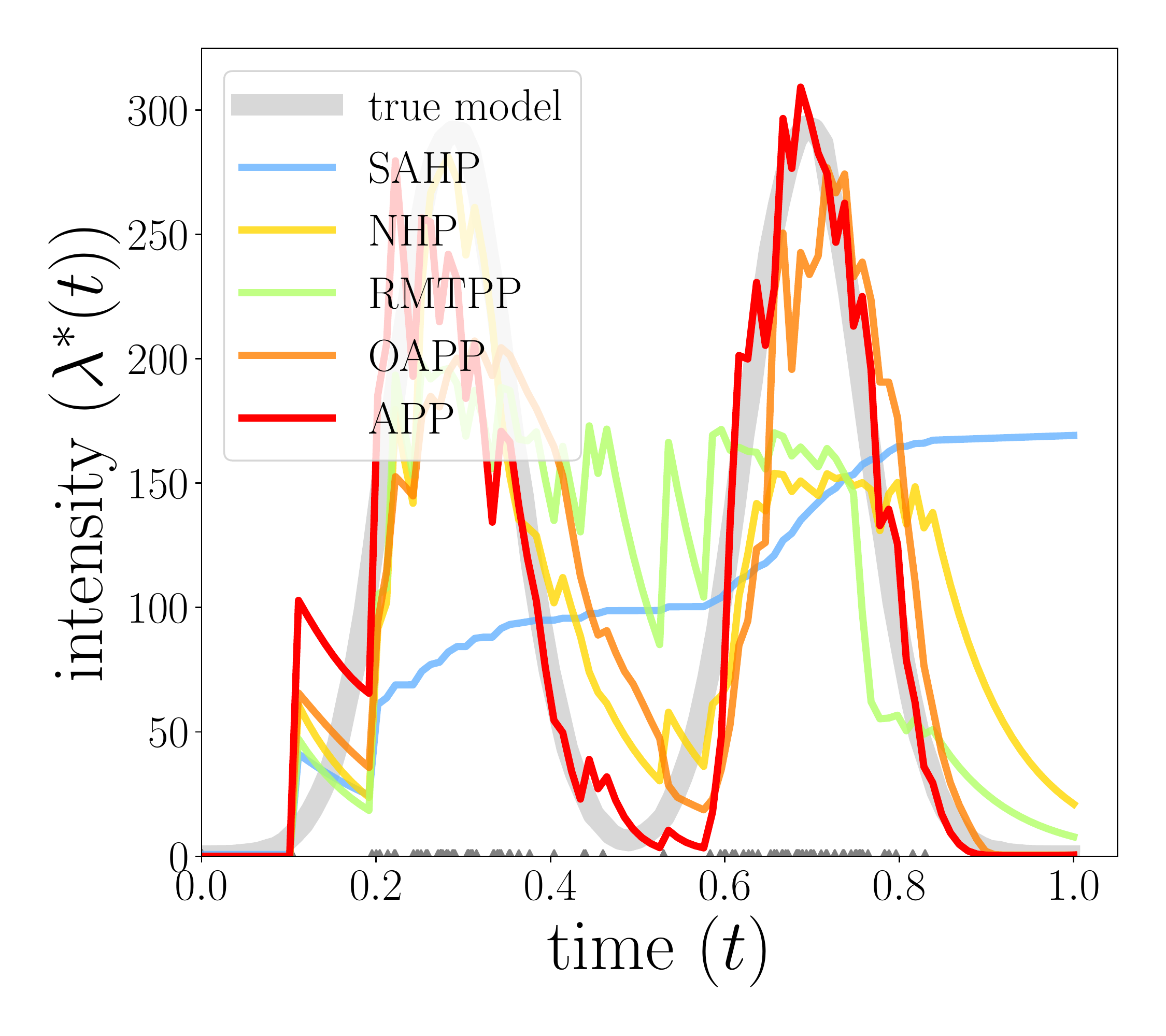}
\caption{non-homogeneous}
% \label{fig:gan-earthquake}
\end{subfigure}
\caption{The conditional intensity function estimated from synthetic datasets. Triangles at the bottom of each panel represent events. The ground truth of conditional intensities is indicated by the grayline.}
\label{fig:sim-res-intensity}
\vspace{-0.15in}
\end{figure*}

\section{Experimental results}
\label{sec:results}

In this section, we first conduct experiments on four synthetic datasets to illustrate our attention-based point process model's effectiveness. Then we test our model on the large-scale Atlanta traffic dataset. We evaluate our model with and without the online attention, and other alternative methods by comparing their log-likelihood and inspecting the recovery of corresponding conditional intensities in both temporal and spatial scenarios. 
Below, we refer to the attention-based point process model as \texttt{APP} and refer to its online version as \texttt{OAPP}.
There are five alternative methods that we are considering in the following experiments:
\begin{enumerate}
\item[(a)] Long-Short Term Memory (\texttt{LSTM}) is a specialized recurrent neural network (RNN) commonly used for sequential data modeling. Here, we feed the event series into the LSTM and obtain the hidden state to summarize the historical information. Then the LSTM can generate the next event via an output layer given the last hidden state \cite{Hochreiter1997}. 

\item[(b)] Recurrent Marked Temporal Point Process (\texttt{RMTPP}) assumes the following form for the conditional intensity function $\lambda^*$ in point processes, denoted as
$
\lambda^*(t) = \exp{\big({\boldsymbol{v}}^{\intercal} \boldsymbol{h}_i+\omega(t-t_i)+b\big)},
$
where the $i$-th hidden state in the RNN $\boldsymbol{h}_i$ is used to represent the influence of historical data up to the most recent event $i$, and $w(t-t_i)$ represents the current influence. The $\boldsymbol{v}, \omega, b$ are trainable parameters \cite{Du2016}.
    
\item[(c)] Neural Hawkes Process (\texttt{NHP}) specifies the conditional intensity function in the temporal point processes using a continuous-time long-short term memory (LSTM), denoted as
$
\lambda^*(t) = f(\boldsymbol{\omega}^\intercal \boldsymbol{h}_i),
$
where the hidden state of the LSTM up to time $t_i$ that represents the influence of historical data, the $f(\cdot)$ is a softplus function which ensures the positive output given any input \cite{Mei2017}. 
    
\item[(d)] Self-Attentive Hawkes Process (\texttt{SAHP}) adopts the self-attention mechanism to model the historical information in the conditional intensity function specified as
$
\lambda^*(t) = \text{softmax}\big(\mu+ \alpha \exp\{\omega(t-t_j)\}\big),
$
where $\mu, \alpha, w$ are computed via three non-linear mappings:
$\mu = \text{softplus}(\boldsymbol{h}W_\mu)$,
$\alpha = \text{tanh}(\boldsymbol{h}W_\alpha)$,
$\omega = \text{softplus}(\boldsymbol{h}W_\omega)$. 
Here $W_\mu, W_\alpha, W_\omega$ are trainable parameters \cite{Zhang2019}. 

\item[(e)] Hawkes Process (\texttt{HP}) is one of the most common model for temporal events data. The conditional intensity function of the Hawkes process is given by 
$
\lambda^*(t) = \mu + \alpha \sum_{t_j < t} \beta \exp\{- \beta (t - t_j)\},
$
where parameters $\mu, \alpha, \beta$ can be estimated via maximizing likelihood \cite{Hawkes1971}.
\end{enumerate}
The experiment setup is as follows. A three-layer neural network in \eqref{eq:score-detail} and three attention heads in \eqref{eq:lambda-prime} are employed in our model. 
Note that using multiple layers in neural networks and multiple heads are critical to achieving our approach's good performance. Each dataset is divided into 80\% training and 20\% testing data. In particular, we use 50\% events in the training data to initialize \texttt{OAPP}; this corresponds to $\eta = 0.5 n$, $n$ is the maximum length of sequences in the dataset. It takes 20 epochs for our model to converge (on a personal laptop, this takes about 12 minutes) for synthetic data. For real traffic data, these numbers increase to 50 epochs and about 45 minutes. To optimize our model, we employ Adam optimizer \cite{Kingma2014} with a learning rate $= 10^{-3}$ while the batch size is 64. The objective is to minimize the negative log-likelihood function derived in \eqref{eq:log-likelihood}. Both our method and alternative approaches are implemented using Python and TensorFlow.

\begin{table}[t]
\caption{Average maximum log-likelihood on synthetic data.}
\label{tab:sim-loglik}
\vspace{-0.1in}
\begin{center}
\begin{small}
\begin{sc}
\resizebox{0.48\textwidth}{!}{%
\begin{tabular}{lcccccr}
\toprule
Data set & SAHP & NHP & RMTPP & APP & OAPP \\
\midrule
Hawkes    & 20.8 & 20.0 & 19.7 & \textbf{21.2} & 21.1  \\
self-correction & 3.5& 5.4& 6.9 & 7.1&\textbf{7.1} \\
non-homo $1$    & 432.4& 445.6&443.1 & 442.3& \textbf{457.0} \\
non-homo $2$    & 364.3& 410.1& 405.1 & \textbf{428.3} & 420.1     \\
\bottomrule
\end{tabular}
}
\end{sc}
\end{small}
\end{center}
\vspace{-0.2in}
\end{table}

%\vspace{-0.1in}
\subsection{Synthetic data}
\label{sec:sim-data}

For the ease of presentation, we compare conditional intensities between approaches using time series data, where each data point is an one-dimensional timestamp. To adapt our \texttt{APP} model to time-only scenarios (in order to compare with the existing methods), we only consider the temporal distance in the score function of \texttt{APP}. i.e., substitute \eqref{eq:score} with $\upsilon_m(x_n, x_i) = \psi_{\theta_m} (t_n - t_i)$. 

The following experiments with synthetic data validate that our \texttt{APP} can capture the temporal pattern of synthetic data generated from conventional generative processes. To quantitatively evaluate each method's performance, we measure the average maximum likelihood on each synthetic dataset. Since the true intensities of these generating processes are known in the synthetic examples, we plot the conditional intensity over time (for one given series of events) to examine if the method can recover the true intensities.

The synthetic data are obtained by the following four generative processes:
(1) \emph{Hawkes process}: the conditional intensity function is given by $\lambda^*(t) = \mu + \alpha \sum_{t_j < t} \beta \exp{- \beta((t-t_j))}$, where $\mu=10$, $\alpha=1$, and $\beta=1$;
(2) \emph{self-correction point process}: the conditional intensity function is given by $\lambda^*(t) = \exp{(\mu t - \sum_{t_i<t}\alpha)}$, where $\mu=10$, $\alpha=1$;
(3) \emph{non-homogeneous Poisson $1$}: The intensity function is given by $\lambda^*(t) = c \cdot\Phi(t-0.5)\cdot U[0,1]$ where $c=100$ is the sample size, the $\Phi(\cdot)$ is the PDF of standard normal distribution, and $U[a,b]$ is uniform distribution between $a$ and $b$;
(4) \emph{non-homogeneous Poisson $2$}: The intensity function is a composition of two normal functions, where $\lambda^*(t) = c_1 \cdot\Phi(6(t-0.35))\cdot U[0,1] + c_2 \cdot\Phi(6(t-0.75))\cdot U[0,1]$, where $c_1=50$, $c_2=50$.
Each synthetic dataset contains 5,000 sequences with an average length of 30, where each data point in the sequence only contains the occurrence time of the event.

Fig.~\ref{fig:sim-res-loglik} summarizes the log-likelihood value of each model versus the training epochs, where each epoch includes 125 batches, and each batch randomly takes 40 sequences as training data. A higher log-likelihood value indicates a better performance of the model. As  Fig.~\ref{fig:sim-res-loglik} and Table~\ref{tab:sim-loglik} show, our \texttt{APP} outperforms other four baseline methods on all four synthetic datasets by converging to larger log-likelihood values. Besides, our \texttt{OAPP} also shows competitive performances, when  only 50\% of events are used in online attention calculation.  

Fig.~\ref{fig:sim-res-intensity} shows the estimated intensities using different methods in contrast to the true latent intensities indicated by the gray lines. We compare the predictive performance of the proposed model fitted to three types of time series models. Our \texttt{APP} can better capture the true conditional intensity function for all four synthetic datasets compared to the other four baseline methods.

% \vspace{-0.1in}
\subsection{Traffic data}
\label{sec:real-data}

This section further evaluates our model on the real Atlanta traffic dataset. 
As a sanity check, we also add two more baselines in the comparison: (i) replacing the tail-up spatial correlation defined in \eqref{eq:alpha} with the Euclidean distance (\texttt{APP+Euclidean}), (ii) removing the exogenous promotion $\mu_1$ defined in \eqref{eq:lambda} (\texttt{APP$_{-\mu_1}$+Tailup}). Below, we refer to our proposed approach as \texttt{APP+Tailup}). 

% The model can be evaluated in the same way via comparing their maximum log-likelihood and prediction accuracy using \eqref{eq:prediction}. Also, we evaluate and visually inspect the conditional intensity for each traffic sensor over time. Moreover, we will also provide an intuitive interpretation for the score and tail-up spatial correlation obtained from the fitted APP model.

\begin{table}[!b]
\vspace{-0.1in}
\caption{Maximum log-likelihood, location-prediction accuracy, and time-prediction error for Atlanta traffic dataset.}
\label{tab:traffic-loglik}
\vspace{-0.1in}
\begin{center}
\begin{small}
\begin{sc}
\resizebox{0.48\textwidth}{!}{%
\begin{tabular}{lccccr}
\toprule
Models & 
\multicolumn{1}{c}{\begin{tabular}[c]{@{}c@{}}$\max \ell$ \\ (time only) \end{tabular}} & 
\multicolumn{1}{c}{\begin{tabular}[c]{@{}c@{}}$\max \ell$ \\ (time \& space) \end{tabular}} & 
\multicolumn{1}{c}{\begin{tabular}[c]{@{}c@{}}location\\accuracy \end{tabular}} & 
\multicolumn{1}{c}{\begin{tabular}[c]{@{}c@{}}time\\MAE \end{tabular}}\\
\midrule
LSTM    & N/A   & N/A   & 18.5\% & 44.0\\
HP      & 339.9     & 307.5    & 8.82\% & 39.1 \\
RMTPP   & 339.2     & 490.1     & 22.0\% & 27.6 \\
NHP     & 324.4    & N/A   & N/A & 24.1 \\
SAHP    & 326.7     & N/A   & N/A & 49.4 \\
APP$_{-\mu_1}$+Tailup & 378.4 & 512.9 & 28.8\% & 17.6 \\
APP + Euclidean    & 392.3     & 570.7     & 30.9\% & 5.5 \\
APP + Tailup   & {\bf 458.5}   & {\bf 636.2}    & {\bf 37.6\%} & \bf{3.7} \\
OAPP + Tailup  & 437.5   & 615.9    & 36.9\% & 3.7 \\
\bottomrule
\end{tabular}
}
\end{sc}
\end{small}
\end{center}
\end{table}

\begin{figure}[!t]
\centering
\begin{subfigure}[h]{0.49\linewidth}
\includegraphics[width=\linewidth]{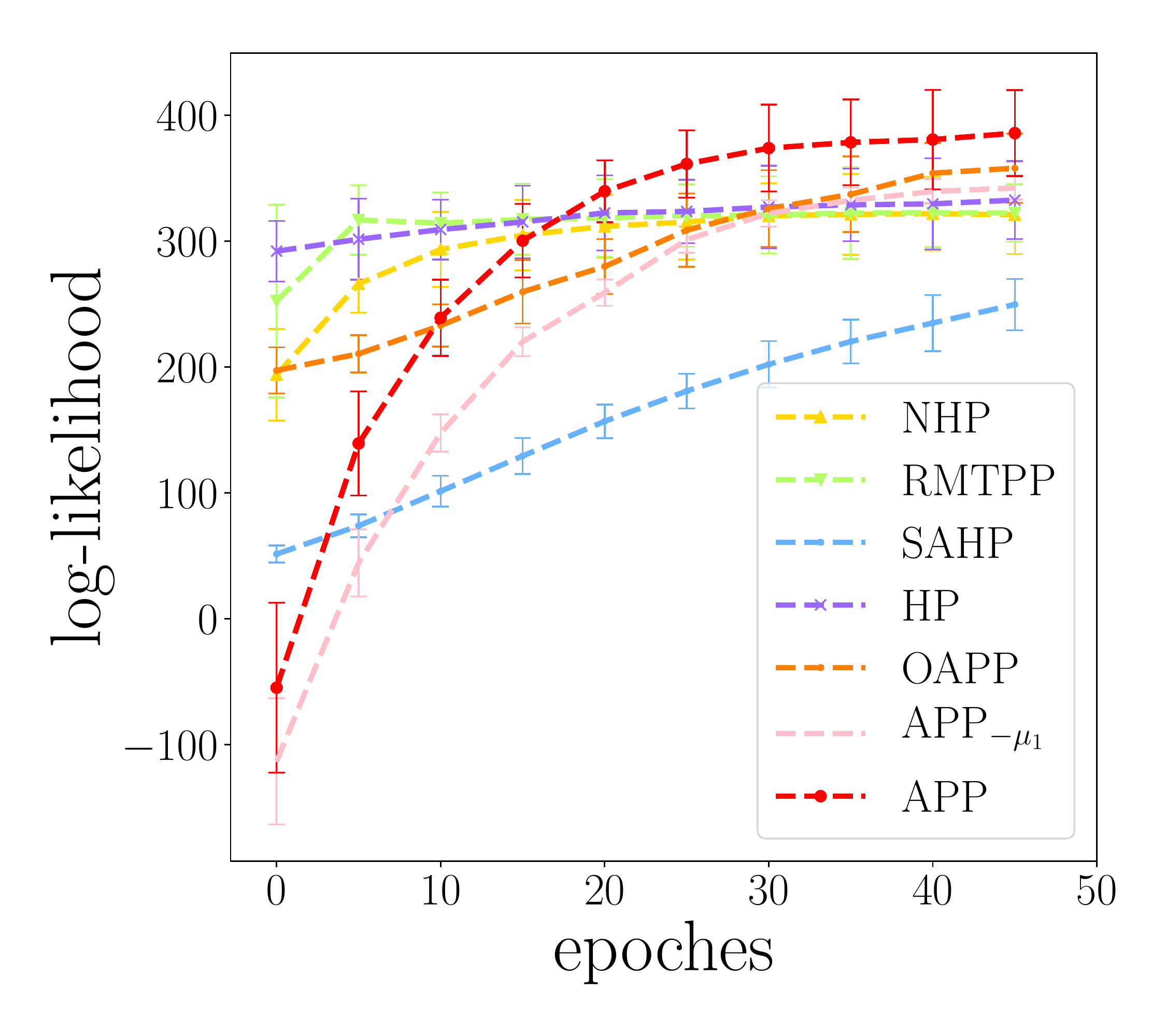}
\caption{time only}
% \label{fig:gan-macys}
\end{subfigure}
\begin{subfigure}[h]{0.49\linewidth}
\includegraphics[width=\linewidth]{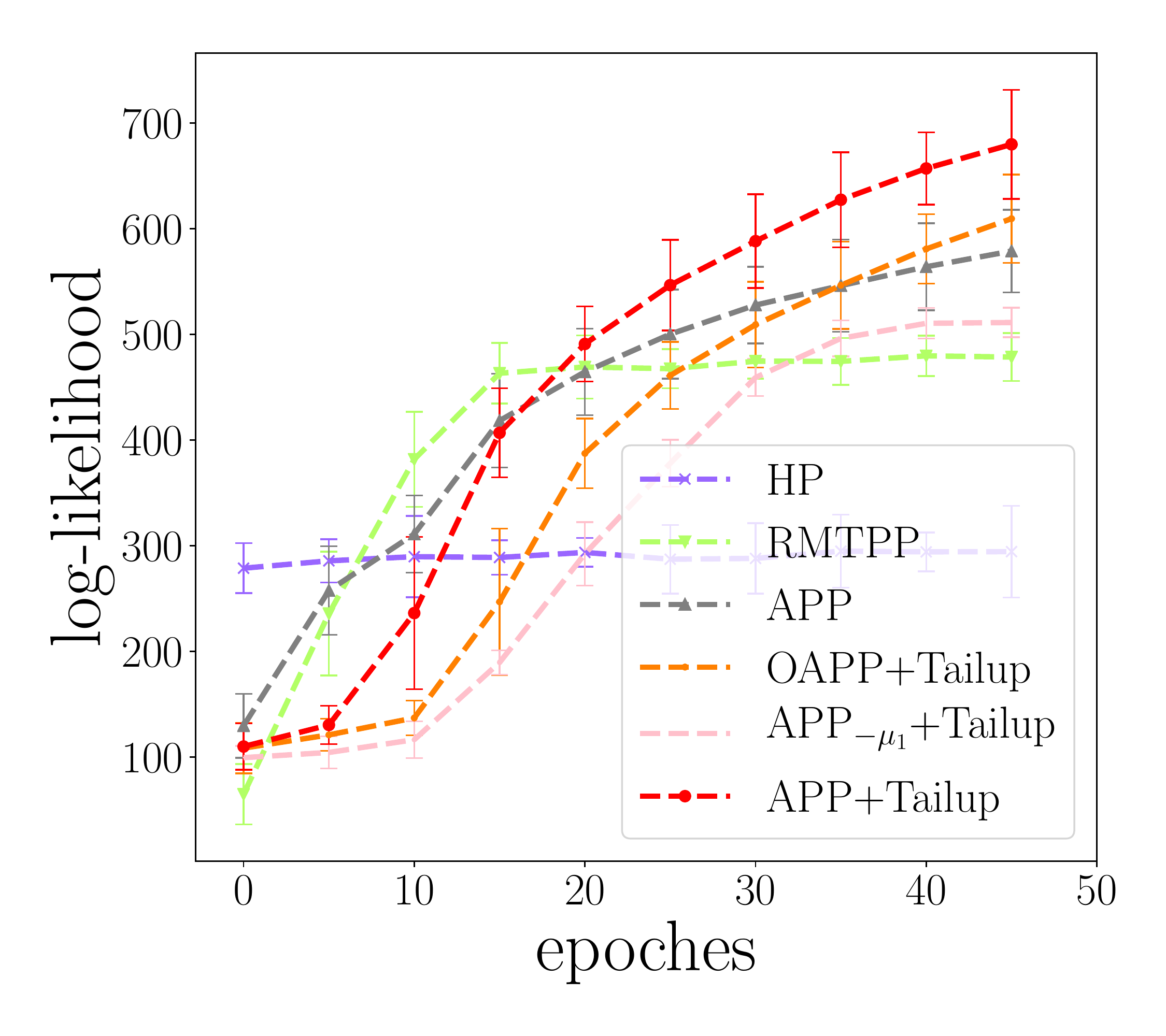}
\caption{space and time}
% \label{fig:gan-earthquake}
\end{subfigure}
\caption{Log-likelihood of Atlanta traffic data versus training epochs with and without considering spatial information.}
\label{fig:traffic-loglik}
\vspace{-0.2in}
\end{figure}

As shown in Fig.~\ref{fig:traffic-loglik}, we report the average log-likelihood per sequence for each method over training epochs. The results show that our method outperforms the competitors by attaining the maximum log-likelihood for both time-only and space-time traffic datasets when the algorithm convergence.
Apart from reporting the maximum log-likelihood of each method, in Table~\ref{tab:traffic-loglik}, we also evaluate the average accuracy for predicting the location of the next event and the mean absolute error (MSE) for predicting the time of the next event, respectively. We use equation \eqref{eq:prediction} for making the prediction and deem it correct if the prediction location is the same as the actual location of the event (recall that the event location is discrete). Then the prediction accuracy is the number of correctly predicted events over the total number of events. We remark that the prediction accuracy, if using random guess, is 7.1\% since there are 14 sensors. Similarly, given observed past events, we can use equation \eqref{eq:prediction} to predict the next event's time. The result shows that our method obtains the highest location accuracy and the lowest time error comparing to other baselines. The experiments also validate that considering the tail-up spatial correlation and the police intervention in our model effectively improves the performance.

\begin{figure*}[!t]
\centering
\begin{subfigure}[h]{0.495\linewidth}
\includegraphics[width=\linewidth]{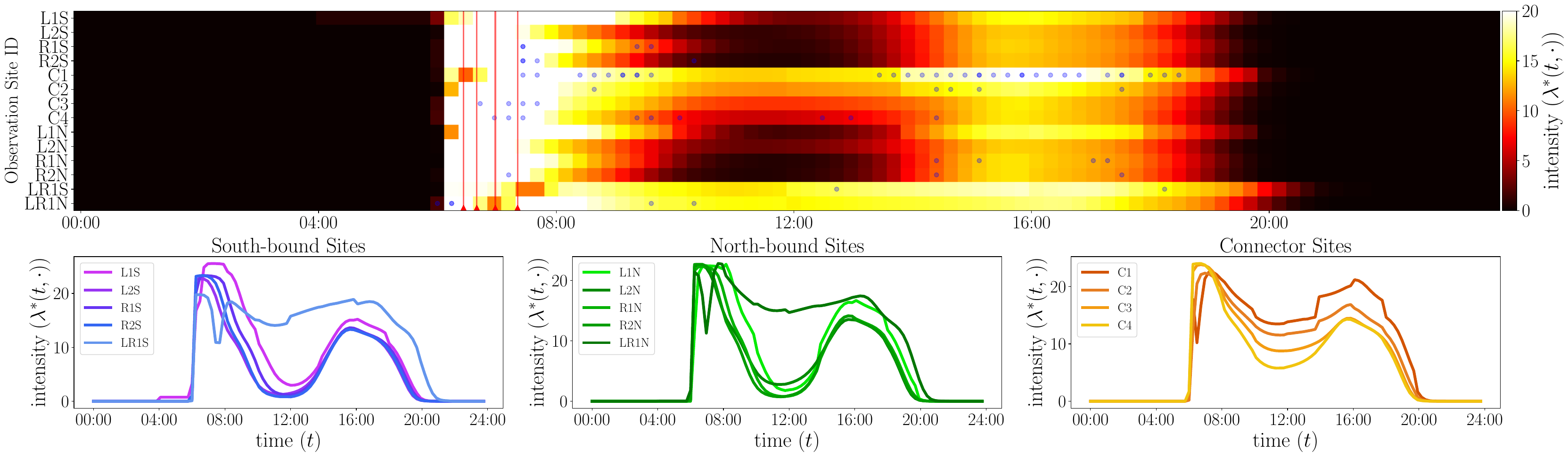}
\caption{Tuesday, May 8th, 2018}
\end{subfigure}
\begin{subfigure}[h]{0.495\linewidth}
\includegraphics[width=\linewidth]{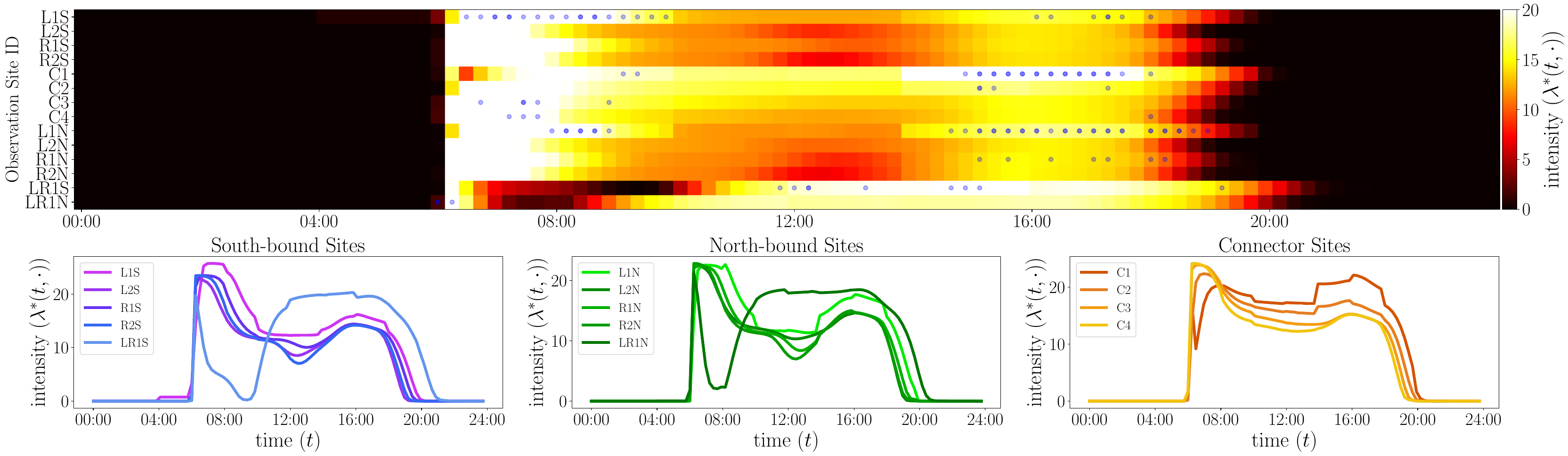}
\caption{Tuesday, April 24th, 2018}
\end{subfigure}
\caption{(a) The heatmap shows the conditional intensities of 14 traffic sensors in a single day. Each row represents a traffic sensor (associated with a unique site ID), each column represents a particular time slot, blue dots correspond to the occurrence of events, and red vertical lines to traffic incidents extracted from 911 calls. The color depth in the heatmap represents the level of intensity. (b) We categorize the conditional intensity into three subplots; the three plots from left to right represent the intensity of five sites on northbound highways, five sites on southbound highways, and four sites on connectors, respectively.}
\label{fig:traffic-res}
\vspace{-0.1in}
\end{figure*}

Moreover, we examine each traffic sensor's conditional intensity computed by the fitted APP on the traffic dataset. As discussed in Section~\ref{sec:intro}, there are 14 traffic sensors installed along two major highways (I-75 and I-85) in Atlanta, as shown in the left of Fig.~\ref{fig:traffic-network}. For better presenting the spatial pattern captured by the model, we select two typical days in Atlanta, which are May 8th, 2018, and April 24th, 2018, respectively. 
Fig.~\ref{fig:traffic-res} (a) shows a clear temporal pattern on a regular weekday (Tuesday, May 8th, 2018), where the intensity of each sensor reaches its pinnacle in both morning (around 7:00) and evening (around 16:00) rush hours; Fig.~\ref{fig:traffic-res} (b) shows the intensities on another weekday (Tuesday, April 24th, 2018). On this day, Atlanta broke a 135-year-old rainfall record when it got 4.16 inches of rain \cite{Nitz2018}. The previous record, set in 1883, was 2.4 inches. As we can see from the figure, the heavy rain and subsequent flood in the city led to an unusual traffic congestion level. Different from the results shown in Fig.~\ref{fig:traffic-res} (a), the traffic congestion level remains at a relatively high level throughout the entire day.

We categorize the traffic sensors into three sub-groups based on their locations and then plot their conditional intensities individually. Note that similar temporal patterns can be found among the traffic sensors in the same sub-group. This can be explained since these sensors are in the same traffic direction and sharing the same traffic flow. The delay can be explained by a so-called ``phantom traffic jam'' phenomenon \cite{Gazis1992}. This situation usually begins when a part of traffic flow slows down slightly, causing the flow behind that part to slow down even more. The effect then spreads out through the traffic network and becomes worse further up the stream. For example, as the sensor \emph{L1S}, \emph{L2S}, \emph{LR1S} show (which are along the southbound of I-75), the peak intensity of the first sensor has an about half an hour delay against the following sensors. Similar phenomenons can also be found among the sensor \emph{L1N}, \emph{L2N}, \emph{LR1N}.

\begin{figure}[!t]
\vspace{-0.1in}
\centering
% \vspace{-0.2in}
\begin{subfigure}[h]{0.48\linewidth}
\includegraphics[width=\linewidth]{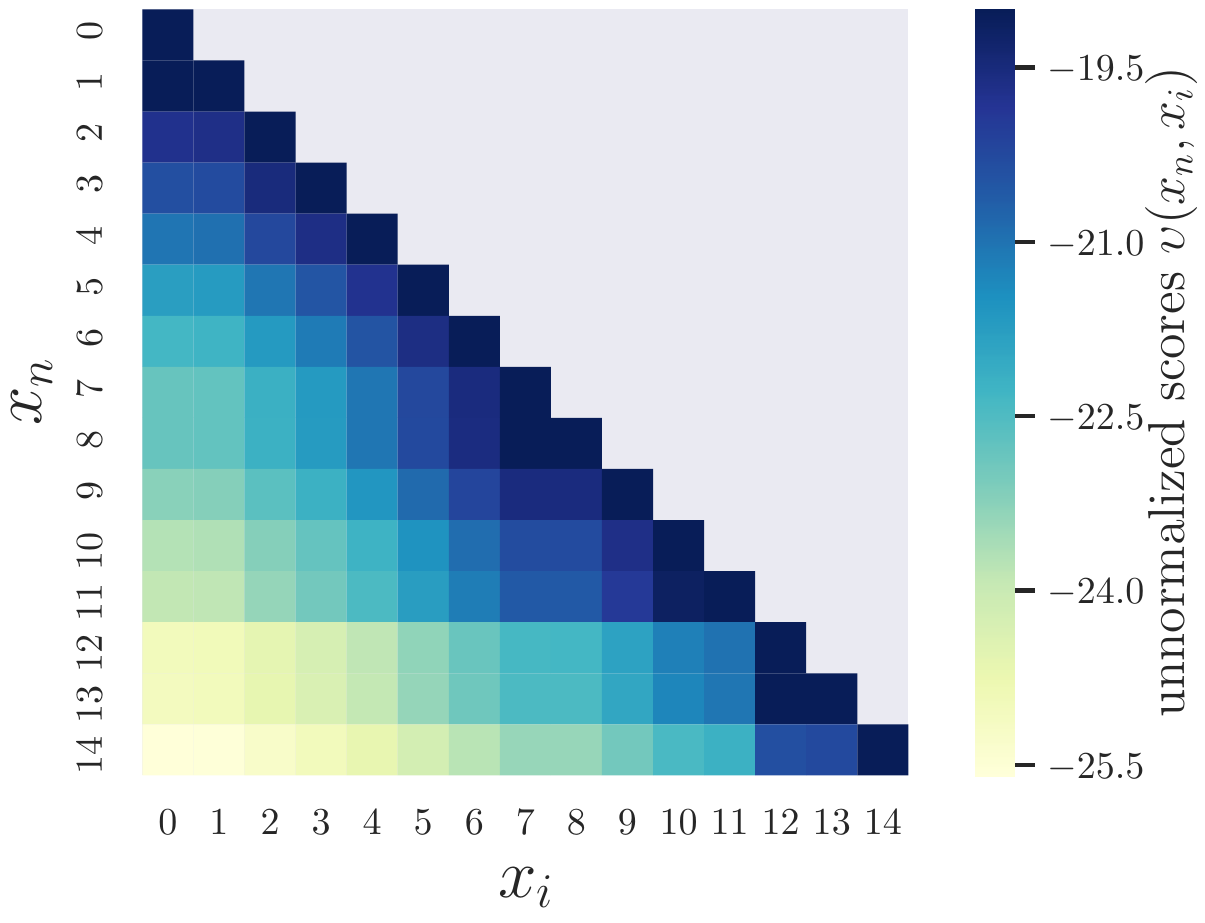}
\caption{synthetic Hawkes}
% \label{fig:gan-macys}
\end{subfigure}
\begin{subfigure}[h]{0.48\linewidth}
\includegraphics[width=\linewidth]{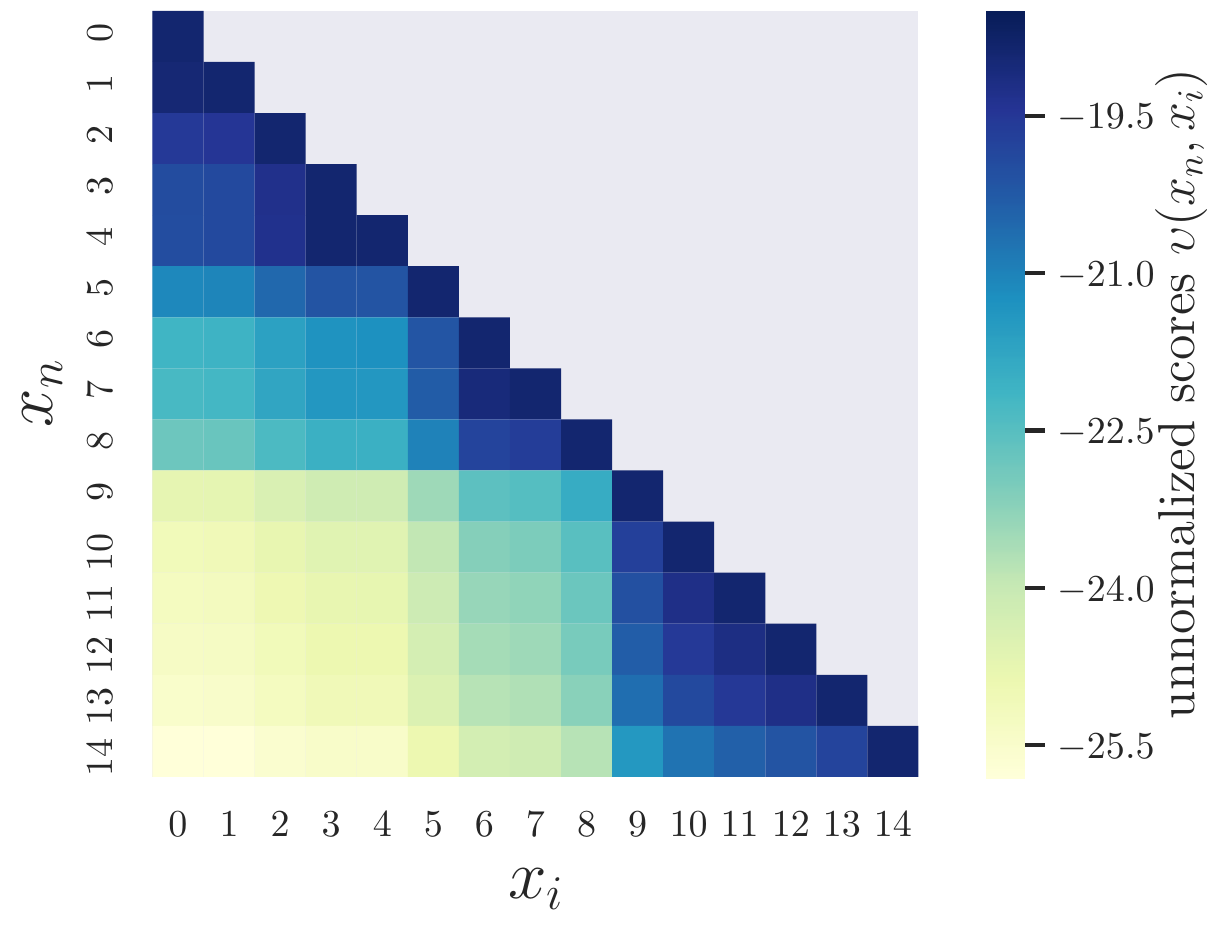}
\caption{real traffic}
% \label{fig:gan-earthquake}
\end{subfigure}
\caption{Visualization of scores (discussed in Section \ref{sec:kernelized-score}) between pairs of events in the sequence for (a) synthetic data generated from a Hawkes process model and (b) traffic data collected from the GDOT. Here $x_n$ represents the current event and $x_i$ represents past events, where $t_i < t_n$. The color of the entry at the $n$-th row and the $i$-th column represents the score $\upsilon(x_n, x_i)$ learned from data using our DAPP model.}
\label{fig:score-visualization}
\end{figure}

\subsection{Score function interpretation}

We now interpret the score function by visualizing the scores of a sequence of 15 congestion events extracted from real data. As shown in Fig.~\ref{fig:score-visualization}, each entry of the heatmap shows the score for one of the event $x_n$ against its historical events $\{x_i\}_{i=1}^{n-1}$. The entry at the $n$-th row and the $i$-th column of the lower triangular matrix represents scores of the event $x_n$ against its past event $x_i$, i.e., $\upsilon(x_n, x_i)$. 

As demonstrated in Fig.~\ref{fig:score-visualization}, our APP can capture complex dependence between events accurately. In particular, Fig.~\ref{fig:score-visualization} (a) shows the scores of events generated from a Hawkes process defined in Section~\ref{sec:sim-data}. Note that the scores for events against their past resemble an exponential decay, reflecting the reality (in this case, the kernel function is exponentially decaying.) We also conduct a similar experiment on the traffic dataset, as shown in Fig.~\ref{fig:score-visualization} (b). Note that there is a ``network community'' structure in that the first nine events pose a much weaker impact on their subsequent events than others, i.e., the first nine scores in the last row are remarkably lower than the other scores. By investigating the data further, we realize that the traffic sensor observes the first nine events on the highway northbound. In contrast, all the other events are observed by sensors installed in the opposite direction. These two sets of traffic sensors are not flow-connected, which explains the score matrix we learned in Fig.~\ref{fig:score-visualization} (b).

\subsection{Tail-up spatial correlation interpretation}

\begin{figure}[!t]
% \vspace{-0.2in}
\centering
\includegraphics[width=.99\linewidth]{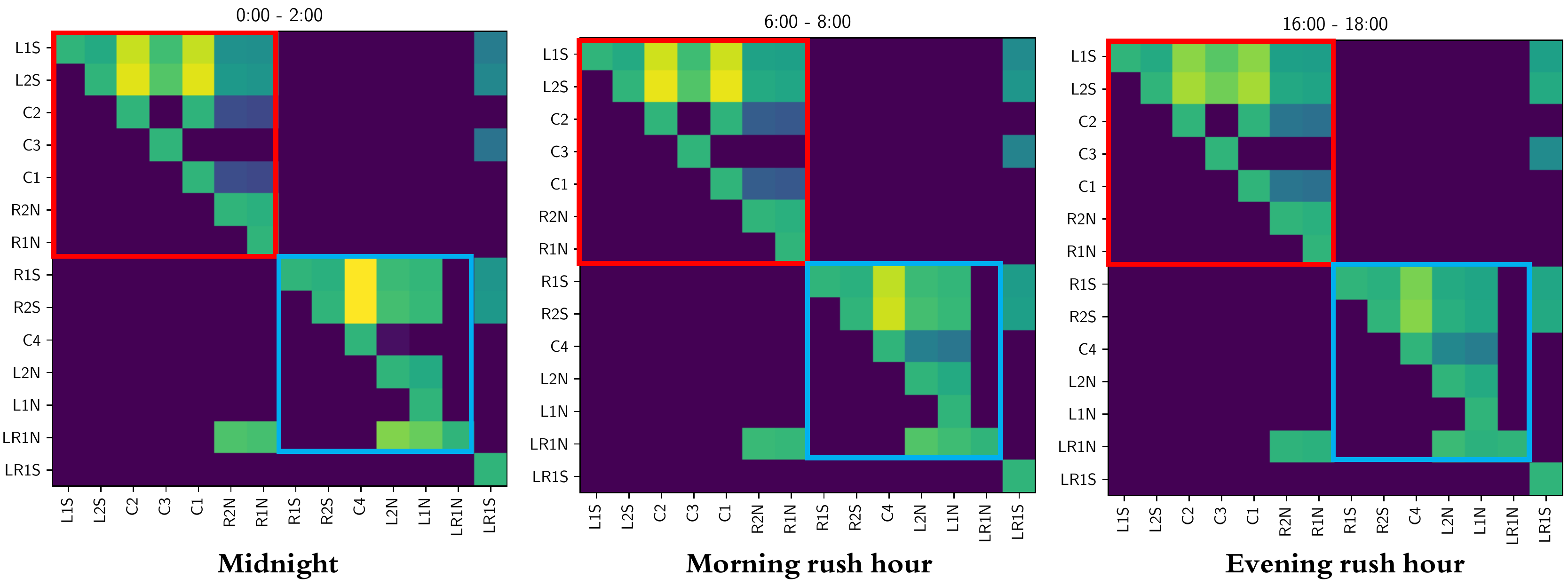}
\caption{Comparison of the spatial correlation between 14 traffic sensors generated by the tail-up model. Each row or column corresponds to one specific traffic sensor. Brighter color indicates a higher correlation. The red and blue boxes include the correlations between traffic sensors located in southbound and northbound of highways, respectively.}
\label{fig:tail-up-covariance}
\vspace{-0.1in}
\end{figure}

The spatial correlation between 14 traffic sensors learned by our tail-up model is highly interpretable. As shown in Fig.~\ref{fig:tail-up-covariance}, we visualize three covariance matrices of 14 traffic sensors corresponding to three different periods: the morning rush hour, the evening rush hour, and the midnight. The spatial structure of these covariance matrices reveals that: (1) There are two ``network communities'' among two sets of traffic sensors, which correspond to the sensors along the southbound and northbound highways. (2) The covariance between the highway connectors (\emph{C1}-\emph{C4}) and the southbound of highways (\emph{L1S}, \emph{L2S}, \emph{R1S}, \emph{R2S}) are different depending on whether it is rush hour or not. These observations confirm the idea that the triggering effect between congestion events can be spatially structured and directional. The highway connectors, in particular, play a vital role in defining such a structure.

% \vspace{-0.1in}
\section{Conclusion}
\label{sec:conclusion}

We developed a novel attention-based point process model for modeling the dynamics of traffic congestion events considering the influence of traffic incidents. In building the model, we combine two data sources: the traffic sensor counts and the police reports (which contain comprehensive records of traffic incidents since police typically are responding to traffic incidents as part of their duty). Our main goal is to model traffic congestion events' self and mutual triggering effect while considering the traffic networks' structures. We adopt the state-of-the-art spatio-temporal point process model, together with the attention models, to achieve our goals. As demonstrated by our experiments, our method achieves a good performance in achieving a higher likelihood of a point process and a higher prediction accuracy compared with previous approaches. By testing on various kinds of point process models, we demonstrate the flexibility of our proposed method. 

%\appendices
%% you can choose not to have a title for an appendix
%% if you want by leaving the argument blank
%\section{}

% \section*{Acknowledgment}


\ifCLASSOPTIONcaptionsoff
  \newpage
\fi



%\vspace{-0.1in}
\bibliographystyle{IEEEtran}
%  argument is your BibTeX string definitions and bibliography database(s)
\bibliography{refs}

\begin{thebibliography}{10}
\providecommand{\url}[1]{#1}
\csname url@samestyle\endcsname
\providecommand{\newblock}{\relax}
\providecommand{\bibinfo}[2]{#2}
\providecommand{\BIBentrySTDinterwordspacing}{\spaceskip=0pt\relax}
\providecommand{\BIBentryALTinterwordstretchfactor}{4}
\providecommand{\BIBentryALTinterwordspacing}{\spaceskip=\fontdimen2\font plus
\BIBentryALTinterwordstretchfactor\fontdimen3\font minus
  \fontdimen4\font\relax}
\providecommand{\BIBforeignlanguage}[2]{{%
\expandafter\ifx\csname l@#1\endcsname\relax
\typeout{** WARNING: IEEEtran.bst: No hyphenation pattern has been}%
\typeout{** loaded for the language `#1'. Using the pattern for}%
\typeout{** the default language instead.}%
\else
\language=\csname l@#1\endcsname
\fi
#2}}
\providecommand{\BIBdecl}{\relax}
\BIBdecl

\bibitem{DavidWickert2020}
\BIBentryALTinterwordspacing
D.~Wickert, ``Is there a fix for {I}-85 traffic?'' Feb 2020. [Online].
  Available:
  \url{\url{https://www.ajc.com/news/transportation/there-fix-for-traffic/FBjOFLgKW7kGphc3itpvnJ/}}
\BIBentrySTDinterwordspacing

\bibitem{Reinhart2017}
A.~Reinhart \emph{et~al.}, ``A review of self-exciting spatio-temporal point
  processes and their applications,'' \emph{Statistical Science}, vol.~33,
  no.~3, pp. 299--318, 2018.

\bibitem{Du2016}
N.~Du, H.~Dai, R.~Trivedi, U.~Upadhyay, M.~Gomez-Rodriguez, and L.~Song,
  ``Recurrent marked temporal point processes: Embedding event history to
  vector,'' in \emph{Proceedings of the 22nd ACM SIGKDD International
  Conference on Knowledge Discovery and Data Mining}, ser. KDD '16.\hskip 1em
  plus 0.5em minus 0.4em\relax New York, NY, USA: Association for Computing
  Machinery, 2016, pp. 1555--1564.

\bibitem{Mei2017}
H.~Mei and J.~M. Eisner, ``The neural hawkes process: A neurally
  self-modulating multivariate point process,'' in \emph{Advances in Neural
  Information Processing Systems 30}, ser. NeurIPS '17.\hskip 1em plus 0.5em
  minus 0.4em\relax Curran Associates, Inc., 2017, pp. 6754--6764.

\bibitem{Li2018}
S.~Li, S.~Xiao, S.~Zhu, N.~Du, Y.~Xie, and L.~Song, ``Learning temporal point
  processes via reinforcement learning,'' in \emph{Proceedings of the 32nd
  International Conference on Neural Information Processing Systems}, ser.
  NeurIPS '18.\hskip 1em plus 0.5em minus 0.4em\relax Red Hook, NY, USA: Curran
  Associates Inc., 2018, pp. 10\,804--10\,814.

\bibitem{Upadhyay2018}
U.~Upadhyay, A.~De, and M.~Gomez~Rodriguez, ``Deep reinforcement learning of
  marked temporal point processes,'' in \emph{Advances in Neural Information
  Processing Systems 31}, ser. NeurIPS '18, S.~Bengio, H.~Wallach,
  H.~Larochelle, K.~Grauman, N.~Cesa-Bianchi, and R.~Garnett, Eds.\hskip 1em
  plus 0.5em minus 0.4em\relax Curran Associates, Inc., 2018, pp. 3168--3178.

\bibitem{Xiao2017A}
S.~Xiao, J.~Yan, X.~Yang, H.~Zha, and S.~M. Chu, ``Modeling the intensity
  function of point process via recurrent neural networks,'' in
  \emph{Proceedings of the Thirty-First AAAI Conference on Artificial
  Intelligence}, ser. AAAI '17.\hskip 1em plus 0.5em minus 0.4em\relax AAAI
  Press, 2017, pp. 1597--1603.

\bibitem{Xiao2017B}
S.~Xiao, M.~Farajtabar, X.~Ye, J.~Yan, L.~Song, and H.~Zha, ``Wasserstein
  learning of deep generative point process models,'' in \emph{Proceedings of
  the 31st International Conference on Neural Information Processing Systems},
  ser. NeurIPS '17.\hskip 1em plus 0.5em minus 0.4em\relax Red Hook, NY, USA:
  Curran Associates Inc., 2017, pp. 3250--3259.

\bibitem{Zhu2019B}
S.~Zhu, H.~S. Yuchi, and Y.~Xie, ``Adversarial anomaly detection for marked
  spatio-temporal streaming data,'' in \emph{ICASSP 2020-2020 IEEE
  International Conference on Acoustics, Speech and Signal Processing
  (ICASSP)}.\hskip 1em plus 0.5em minus 0.4em\relax IEEE, 2020, pp. 8921--8925.

\bibitem{rizoiu2017}
M.-A. Rizoiu, L.~Xie, S.~Sanner, M.~Cebrian, H.~Yu, and P.~Van~Hentenryck,
  ``Expecting to be hip: Hawkes intensity processes for social media
  popularity,'' in \emph{Proceedings of the 26th International Conference on
  World Wide Web}, ser. WWW '17, 2017, pp. 735--744.

\bibitem{Luong2015}
T.~Luong, H.~Pham, and C.~D. Manning, ``Effective approaches to attention-based
  neural machine translation,'' in \emph{Proceedings of the 2015 Conference on
  Empirical Methods in Natural Language Processing}, ser. EMNLP '15.\hskip 1em
  plus 0.5em minus 0.4em\relax Lisbon, Portugal: Association for Computational
  Linguistics, Sep. 2015, pp. 1412--1421.

\bibitem{Vaswani2017}
A.~Vaswani, N.~Shazeer, N.~Parmar, J.~Uszkoreit, L.~Jones, A.~N. Gomez,
  L.~Kaiser, and I.~Polosukhin, ``Attention is all you need,'' in
  \emph{Advances in Neural Information Processing Systems 30}, ser. NeurIPS
  '17.\hskip 1em plus 0.5em minus 0.4em\relax Curran Associates, Inc., 2017,
  pp. 5998--6008.

\bibitem{Abadi2015}
A.~{Abadi}, T.~{Rajabioun}, and P.~A. {Ioannou}, ``Traffic flow prediction for
  road transportation networks with limited traffic data,'' \emph{IEEE
  Transactions on Intelligent Transportation Systems}, vol.~16, no.~2, pp.
  653--662, April 2015.

\bibitem{Lv2015}
Y.~{Lv}, Y.~{Duan}, W.~{Kang}, Z.~{Li}, and F.~{Wang}, ``Traffic flow
  prediction with big data: A deep learning approach,'' \emph{IEEE Transactions
  on Intelligent Transportation Systems}, vol.~16, no.~2, pp. 865--873, April
  2015.

\bibitem{ma2017}
X.~Ma, Z.~Dai, Z.~He, J.~Ma, Y.~Wang, and Y.~Wang, ``Learning traffic as
  images: A deep convolutional neural network for large-scale transportation
  network speed prediction,'' \emph{Sensors}, vol.~17, no.~4, p. 818, 2017.

\bibitem{cui2018}
Z.~Cui, R.~Ke, Z.~Pu, and Y.~Wang, ``Deep bidirectional and unidirectional
  {LSTM} recurrent neural network for network-wide traffic speed prediction,''
  \emph{arXiv preprint arXiv:1801.02143}, 2018.

\bibitem{Liao2018}
B.~Liao, J.~Zhang, C.~Wu, D.~McIlwraith, T.~Chen, S.~Yang, Y.~Guo, and F.~Wu,
  ``Deep sequence learning with auxiliary information for traffic prediction,''
  in \emph{Proceedings of the 24th ACM SIGKDD International Conference on
  Knowledge Discovery \& Data Mining}, ser. KDD '18.\hskip 1em plus 0.5em minus
  0.4em\relax New York, NY, USA: Association for Computing Machinery, 2018, pp.
  537--546.

\bibitem{Yuan2018}
Z.~Yuan, X.~Zhou, and T.~Yang, ``{Hetero-ConvLSTM}: A deep learning approach to
  traffic accident prediction on heterogeneous spatio-temporal data,'' in
  \emph{Proceedings of the 24th ACM SIGKDD International Conference on
  Knowledge Discovery \& Data Mining}, ser. KDD '18.\hskip 1em plus 0.5em minus
  0.4em\relax New York, NY, USA: Association for Computing Machinery, 2018, pp.
  984--992.

\bibitem{Gu2019}
Y.~{Gu}, W.~{Lu}, X.~{Xu}, L.~{Qin}, Z.~{Shao}, and H.~{Zhang}, ``An improved
  {B}ayesian combination model for short-term traffic prediction with deep
  learning,'' \emph{IEEE Transactions on Intelligent Transportation Systems},
  vol.~21, no.~3, pp. 1332--1342, 2019.

\bibitem{Pan2019}
Z.~Pan, Y.~Liang, W.~Wang, Y.~Yu, Y.~Zheng, and J.~Zhang, ``Urban traffic
  prediction from spatio-temporal data using deep meta learning,'' in
  \emph{Proceedings of the 25th ACM SIGKDD International Conference on
  Knowledge Discovery \& Data Mining}, ser. KDD '19.\hskip 1em plus 0.5em minus
  0.4em\relax New York, NY, USA: Association for Computing Machinery, 2019, pp.
  1720--1730.

\bibitem{Zheng2019}
C.~Zheng, X.~Fan, C.~Wang, and J.~Qi, ``{GMAN}: A graph multi-attention network
  for traffic prediction,'' in \emph{Proceedings of the AAAI Conference on
  Artificial Intelligence}, ser. AAAI '20, vol.~34, no.~01, 2020, pp.
  1234--1241.

\bibitem{ZhuL2019}
L.~{Zhu}, F.~R. {Yu}, Y.~{Wang}, B.~{Ning}, and T.~{Tang}, ``Big data analytics
  in intelligent transportation systems: A survey,'' \emph{IEEE Transactions on
  Intelligent Transportation Systems}, vol.~20, no.~1, pp. 383--398, Jan 2019.

\bibitem{wilson2001}
R.~E. Wilson, ``An analysis of gipps's car-following model of highway
  traffic,'' \emph{IMA journal of applied mathematics}, vol.~66, no.~5, pp.
  509--537, 2001.

\bibitem{zeroual2017}
A.~Zeroual, F.~Harrou, Y.~Sun, and N.~Messai, ``Monitoring road traffic
  congestion using a macroscopic traffic model and a statistical monitoring
  scheme,'' \emph{Sustainable cities and society}, vol.~35, pp. 494--510, 2017.

\bibitem{sole2016}
A.~Sol{\'e}-Ribalta, S.~G{\'o}mez, and A.~Arenas, ``A model to identify urban
  traffic congestion hotspots in complex networks,'' \emph{Royal Society open
  science}, vol.~3, no.~10, p. 160098, 2016.

\bibitem{Hawkes1971}
A.~G. Hawkes, ``Spectra of some self-exciting and mutually exciting point
  processes,'' \emph{Biometrika}, vol.~58, no.~1, pp. 83--90, 1971.

\bibitem{Gomez2010}
M.~Gomez~Rodriguez, J.~Leskovec, and A.~Krause, ``Inferring networks of
  diffusion and influence,'' in \emph{Proceedings of the 16th ACM SIGKDD
  International Conference on Knowledge Discovery and Data Mining}, ser. KDD
  '10.\hskip 1em plus 0.5em minus 0.4em\relax New York, NY, USA: Association
  for Computing Machinery, 2010, pp. 1019--1028.

\bibitem{Yuan2019}
B.~Yuan, H.~Li, A.~L. Bertozzi, P.~J. Brantingham, and M.~A. Porter,
  ``Multivariate spatiotemporal hawkes processes and network reconstruction,''
  \emph{SIAM Journal on Mathematics of Data Science}, vol.~1, no.~2, pp.
  356--382, 2019.

\bibitem{Zhu2019A}
S.~Zhu and Y.~Xie, ``Spatial-temporal-textual point processes with applications
  in crime linkage detection,'' \emph{arXiv preprint arXiv:1902.00440}, 2019.

\bibitem{Omi2019}
T.~Omi, N.~Ueda, and K.~Aihara, ``Fully neural network based model for general
  temporal point processes,'' in \emph{Advances in Neural Information
  Processing Systems 32}, ser. NeurIPS '19.\hskip 1em plus 0.5em minus
  0.4em\relax Curran Associates, Inc., 2019, pp. 2120--2129.

\bibitem{Zhang2019}
Q.~Zhang, A.~Lipani, O.~Kirnap, and E.~Yilmaz, ``Self-attentive {H}awkes
  process,'' in \emph{Proceedings of the 37th International Conference on
  Machine Learning}, ser. Proceedings of Machine Learning Research, vol.
  119.\hskip 1em plus 0.5em minus 0.4em\relax PMLR, 13--18 Jul 2020, pp.
  11\,183--11\,193.

\bibitem{GDOT}
\BIBentryALTinterwordspacing
{Georgia Department of Transportation}, ``Traffic analysis and data application
  (tada).'' [Online]. Available: \url{http://www.dot.ga.gov/DS/Data}
\BIBentrySTDinterwordspacing

\bibitem{treiber2013traffic}
M.~Treiber and A.~Kesting, ``Traffic flow dynamics,'' \emph{Traffic Flow
  Dynamics: Data, Models and Simulation}, 2013.

\bibitem{Zhu2019C}
S.~Zhu and Y.~Xie, ``Crime event embedding with unsupervised feature
  selection,'' in \emph{2019 IEEE International Conference on Acoustics, Speech
  and Signal Processing (ICASSP)}, May 2019, pp. 3922--3926.

\bibitem{OpenStreetMap}
\BIBentryALTinterwordspacing
{OpenStreetMap contributors}, ``Openstreetmap,'' 2017. [Online]. Available:
  \url{\url{https://www.openstreetmap.org}}
\BIBentrySTDinterwordspacing

\bibitem{Barry1996}
R.~P. Barry and J.~M.~V. Hoef, ``Blackbox {Kriging}: Spatial prediction without
  specifying variogram models,'' \emph{Journal of Agricultural, Biological, and
  Environmental Statistics}, vol.~1, no.~3, pp. 297--322, 1996.

\bibitem{Garreta2010}
V.~Garreta, P.~Monestiez, and J.~M. Ver~Hoef, ``Spatial modelling and
  prediction on river networks: Up model, down model or hybrid?''
  \emph{Environmetrics}, vol.~21, no.~5, pp. 439--456, 2010.

\bibitem{Hoef2006}
J.~M.~V. Hoef, E.~Peterson, and D.~Theobald, ``Spatial statistical models that
  use flow and stream distance,'' \emph{Environmental and Ecological
  Statistics}, vol.~13, no.~4, pp. 449--464, Dec 2006.

\bibitem{Hoef2010}
J.~M.~V. Hoef and E.~E. Peterson, ``A moving average approach for spatial
  statistical models of stream networks,'' \emph{Journal of the American
  Statistical Association}, vol. 105, no. 489, pp. 6--18, 2010.

\bibitem{Chen2017}
J.~Chen, S.-H. Kim, and Y.~Xie, ``$\textsf{S}^3t$: An efficient score-statistic
  for spatio-temporal surveillance,'' \emph{forthcoming in Sequential Analysis:
  Design Methods and Applications}, 2020.

\bibitem{Cressie2006}
N.~Cressie, J.~Frey, B.~Harch, and M.~Smith, ``Spatial prediction on a river
  network,'' \emph{Journal of Agricultural, Biological, and Environmental
  Statistics}, vol.~11, no.~2, p. 127, Jun 2006.

\bibitem{Peterson2007}
E.~E. Peterson, D.~M. Theobald, and J.~M. Ver~Hoef, ``Geostatistical modelling
  on stream networks: developing valid covariance matrices based on hydrologic
  distance and stream flow,'' \emph{Freshwater Biology}, vol.~52, no.~2, pp.
  267--279, 2007.

\bibitem{Hochreiter1997}
S.~Hochreiter and J.~Schmidhuber, ``Long short-term memory,'' \emph{Neural
  Comput.}, vol.~9, no.~8, pp. 1735--1780, Nov. 1997.

\bibitem{Kingma2014}
D.~P. Kingma and J.~Ba, ``Adam: {A} method for stochastic optimization,'' in
  \emph{3rd International Conference on Learning Representations, {ICLR} 2015,
  San Diego, CA, USA, May 7-9, 2015, Conference Track Proceedings}, Y.~Bengio
  and Y.~LeCun, Eds., 2015.

\bibitem{Nitz2018}
\BIBentryALTinterwordspacing
B.~Nitz, ``Atlanta breaks 135-year-old rainfall record,'' April 2018. [Online].
  Available:
  \url{\url{https://www.wsbtv.com/news/local/wet-roads-could-make-for-messy-commute/737464010/}}
\BIBentrySTDinterwordspacing

\bibitem{Gazis1992}
D.~C. Gazis and R.~Herman, ``The moving and “phantom” bottlenecks,''
  \emph{Transportation Science}, vol.~26, no.~3, pp. 223--229, 1992.

\end{thebibliography}
\end{document}